\documentclass[acmtog,screen,authorversion]{acmart}

\acmSubmissionID{198}
\AtBeginDocument{%
  \providecommand\BibTeX{{%
    \normalfont B\kern-0.5em{\scshape i\kern-0.25em b}\kern-0.8em\TeX}}}

\setcopyright{acmlicensed}
\acmJournal{TOG}
\acmYear{2023}
\acmVolume{42}
\acmNumber{4}
\acmArticle{}
\acmMonth{8}
\acmPrice{15.00}
\acmDOI{10.1145/3592103}

\usepackage{my}

\begin{document}

\title{Locally Attentional SDF Diffusion for Controllable 3D Shape Generation}

\author{Xin-Yang Zheng}
\email{zxy20@mails.tsinghua.edu.cn}
\affiliation{%
  \institution{Tsinghua University}
  \city{Beijing}
  \country{P.~R.~China}
}
\author{Hao Pan}
\email{haopan@microsoft.com}
\affiliation{%
  \institution{Microsoft Research Asia}
  \city{Beijing}
  \country{P.~R.~China}
}
\author{Peng-Shuai Wang}
\email{wangps@hotmail.com}
\affiliation{%
  \institution{Peking University}
  \city{Beijing}
  \country{P.~R.~China}
}
\author{Xin Tong}
\email{xtong@microsoft.com}
\affiliation{%
  \institution{Microsoft Research Asia}
  \city{Beijing}
  \country{P.~R.~China}
}
\author{Yang Liu}
\email{yangliu@microsoft.com}
\affiliation{%
  \institution{Microsoft Research Asia}
  \city{Beijing}
  \country{P.~R.~China}
}
\author{Heung-Yeung Shum}
\email{msraharry@hotmail.com}
\affiliation{%
  \institution{Tsinghua University \& International Digital Economy Academy}
  \city{Beijing}
  \country{P.~R.~China}
}
\authorsaddresses{Xin-Yang Zheng (Work done during internship at Microsoft Research Asia), zxy20@mails.tsinghua.edu.cn; Hao Pan, haopan@microsoft.com;  Peng-Shuai Wang, wangps@hotmail.com; Xin Tong, xtong@microsoft.com; Yang Liu (corresponding author), yangliu@microsoft.com; Heung-Yeung Shum, msraharry@hotmail.com.}

\renewcommand{\shortauthors}{Zheng, et al.}

\begin{abstract}
  Although the recent rapid evolution of 3D generative neural networks greatly improves 3D shape generation, it is still not convenient for ordinary users to create 3D shapes and control the local geometry of generated shapes. To address these challenges, we propose a diffusion-based 3D generation framework --- \emph{locally attentional SDF diffusion}, to model plausible 3D shapes, via 2D sketch image input. Our method is built on a two-stage diffusion model. The first stage, named \emph{occupancy-diffusion}, aims to generate a low-resolution occupancy field to approximate the shape shell. The second stage, named \emph{SDF-diffusion}, synthesizes a high-resolution signed distance field within the occupied voxels determined by the first stage to extract fine geometry. Our model is empowered by a novel \emph{view-aware local attention} mechanism for image-conditioned shape generation, which takes advantage of 2D image patch features to guide 3D voxel feature learning, greatly improving local controllability and model generalizability.
Through extensive experiments in sketch-conditioned and category-conditioned 3D shape generation tasks, we validate and demonstrate the ability of our method to provide plausible and diverse 3D shapes, as well as its superior controllability and generalizability over existing work.
\end{abstract}

\ccsdesc[500]{Computing methodologies~Shape modeling}
\ccsdesc[500]{Computing methodologies~Neural networks}

\keywords{3D shape generation, diffusion model, sketch-conditioned, local attention}

\begin{teaserfigure}
  \centering
\begin{overpic}[width=\textwidth]{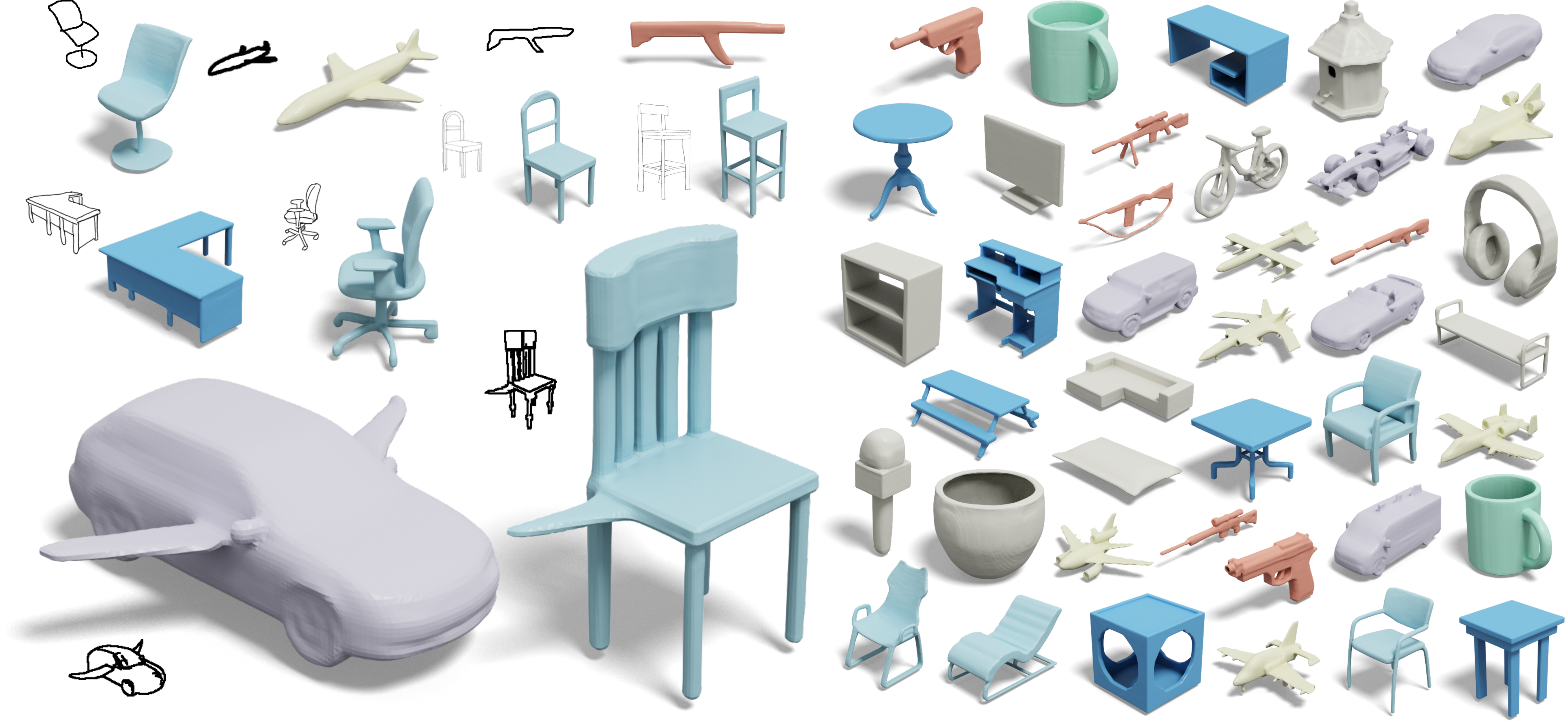}
\end{overpic}
\caption{Plausible and diverse shapes generated by our LAS-Diffusion model. \textbf{Left}: Our sketch-conditioned model supports freehand sketches (top), and is able to generate novel 3D shapes such as a flying car and a chair with a wing (bottom), which have not been seen in the training data. \textbf{Right}: a shape gallery generated by our category-conditioned model.}
\label{fig:teaser}
\end{teaserfigure}

\maketitle
\section{Introduction} \label{sec:intro}

Easily creating 3D shapes to fit human's fabulous imaginations and match the designer's creative ideas is one of the ultimate goals in computer graphics. The rapid development of generative neural networks, such as generative adversarial networks (GAN)~\cite{goodfellow2014generative}, diffusion models~\cite{ho2020denoising}, autoregressive networks~\cite{van2016pixel}, and flow-based models~\cite{rezende2015variational}, achieves great progress in text, image, and video generation. These techniques have been adopted for generating 3D shapes with different kinds of 3D representations and greatly reduce the workload of 3D generation. However, there exists a large quality gap between the synthesized shapes and the dataset the generator was trained on. Moreover, existing approaches lack intuitive control and convenient ways to control the shape generation process to satisfy users' intentions.

For the quality gap, our key observation is that it is due to two factors. First, the underlying 3D representation affects the generated geometry quality. Previous methods focus mainly on generating 3D shapes with discrete point cloud or voxel representations. The discretization error caused by the limited output resolution degrades the output quality. Furthermore, an explicit conversion step is usually required to convert the discrete results into continuous shape geometry, which is fragile to reconstruct high-quality geometry. Second, the capability of the chosen generative technique may be limited for modeling 3D shapes with complex structures. As observed in image synthesis~\cite{dhariwal2021diffusion}, GAN-based generation tends to have less diversity than diffusion models. We also found that 3D shapes with complex structures are difficult to be learned and generated by 3D GANs.

For intuitive control, many existing 3D shape generation works focus on unconditional shape generation.  It becomes difficult for normal users to embed their creative ideas into the generation process. Using text as conditions to guide 3D generation~\cite{chen2018text2shape,Sanghi_2022_CVPR} is a promising way to loop humans in content generation. Despite rapid progress in text-to-image and text-to-3D development, users still need to spend considerable time in prompt engineering to seek satisfying results, and the diversity and the amount of paired 3D-text datasets for training and fine-tuning deep generative techniques are very limited. On the contrary, 2D sketching is a natural interface for people with diverse backgrounds to depict, explore, and exchange creative ideas~\cite{eitz2012humans,olsen2009sketch}, without suffering language barriers. However, existing sketch-based generation techniques do not provide local controllability and have limited generalizability to unseen shapes, as they often encode a whole sketch as a global feature for use.  

In this work, we propose a novel diffusion-based 3D shape generation approach to address the above challenges. To overcome the quality gap, our approach utilizes the SDF representation and the powerful diffusion model for 3D shape generation. The main challenge is that a na\"{i}ve high-resolution SDF diffusion in the 3D space is costly due to high memory consumption and heavy computation. Thus, we perform two-stage diffusion to minimize memory and computational cost. The first stage is called \emph{occupancy-diffusion}, which transforms random noises to a coarse occupancy field to model the shell of a 3D shape surface; the second step, called \emph{SDF-diffusion}, plays an upsampler role that generates a high-resolution SDF within the occupied region determined in the first stage. For local control, our method takes sketches as input to achieve local controllability and better generalizability in 3D shape generation. To this end, we introduce a \emph{view-aware local attention} mechanism that takes 2D sketches as input and interacts with the 3D diffusion models with learned attention. We name our approach --- \emph{locally attentional SDF diffusion}, dubbed \emph{LAS-Diffusion}.

Our model produces good-quality shapes to match the user's input sketch and is robust to both synthetic sketches extracted from 2D images and free-hand sketches. We validate our model design via extensive evaluations and demonstrate the superiority of our approach over other existing shape synthesis works, in terms of local controllability and model generalizability for sketch-conditioned shape generation, shape quality and diversity for category-conditioned shape generation. Our code and pre-trained models are available at: \url{https://zhengxinyang.github.io/projects/LAS-Diffusion.html}.

\section{Related Work} \label{sec:related}

\paragraph{Shape representations in 3D generation}
Early 3D generation works adopt low-resolution occupancy fields~\cite{wu2016learning}, and fixed-number points~\cite{achlioptas2018learning} as shape representations. Their representation ability is limited by their discrete nature, and further refinements~\cite{chen2021decor,hui2020progressive} are needed. Polygonal meshes are also used for 3D generation~\cite{wang2018pixel2mesh,khalid2022text,nash2020polygen,gao2022get3d} as they are suitable for many downstream tasks. Recently, implicit representations such as implicit occupancy fields, signed distance functions (SDF) and neural radiance fields (NeRF) are preferable for 3D shape generation~\cite{jiang2017hierarchical,chen2019learning,kleineberg2020adversarial,schwarz2020graf}, due to their great capability in modeling varied shape geometry, even appearance. In our work, we choose voxel-based SDFs as our 3D representation and enable high-resolution SDF generation via two-stage diffusion.  In the following, we briefly review the most relevant works to our approach.

\paragraph{GAN-based 3D generation} The works of \cite{chen2019learning,ibing20213d} trained an implicit autoencoder that encodes shape collection in a latent space, then applied latent-GAN to sample latent codes and decode them as implicit occupancy fields. 
Kleineberg \etal~\shortcite{kleineberg2020adversarial} and Zheng \etal~\shortcite{zheng2022sdfstylegan} directly discriminated the 3D output, and the latter method combined local and global discriminators to improve shape quality.
Some recent methods~\cite{niemeyer2021giraffe,chan2021pi,gao2022get3d,chan2022efficient,deng2022gram,or2022stylesdf} directly use adversarial losses built on image rendering to guide network training, without 3D supervision.

\paragraph{Autoregressive-based 3D generation} Ibing \etal~\shortcite{ibing2021octree} sequentially generated an octree structure that hierarchically represents the 3D occupancy.  The work of AutoSDF~\cite{autosdf2022} and ShapeFormer~\cite{yan2022shapeformer} used VQ-VAE~\cite{van2017neural} or its variants on implicit functions to encode regular voxel patches into a latent space, then learn a transformer-based autoregressive model over the latent space. 
Zhang \etal~\shortcite{zhang20223dilg} avoided encoding empty voxels and defined latents on irregular grids, further improving the power of autoregressive-based 3D models.

\paragraph{Diffusion-based 3D generation} The success of diffusion models in image generation inspires many 3D point cloud generation work~\cite{cai2020learning,luo2021diffusion,zhou20213d,lyu2021conditional,kong2022diffusion,zeng2022lion}. However, additional and nontrivial efforts are needed to convert point clouds to continuous shapes.  To directly leverage SDF representation,  Hui \etal~\shortcite{hui2022neural} developed diffusion-based generators to produce coarse and detailed coefficient volumes, which can be transformed back into truncated SDFs. Latent diffusion models for SDF and occupancy generation are also explored in recent concurrent work~\cite{li2022diffusion,cheng2022sdfusion,chou2022diffusionsdf,nam20223d}: an SDF autoencoder is first trained to build the latent space, similar to latent-GAN; and a diffusion model is trained to generate the latent code that can be transformed to SDF by the pre-trained decoder.
Shue \etal~\shortcite{shue20223d} used triplane features to further improve latent expressiveness and allowed using high-resolution occupancy fields for training.
Unlike these approaches, our diffusion model operates on the 3D SDF space directly to easily incorporate local features from conditional inputs to achieve better controllability and generalizability.

\paragraph{Conditional 3D generation} Various input conditions, such as texts, images, coarse voxels, sparse points, and bounding volumes, have been used for 3D generation to assist content creation and improve downstream tasks like voxel super-resolution,  shape reconstruction, and completion~\cite{cheng2022cross,chen2018text2shape,fu2022shapecrafter}.
Some recent works \cite{Sanghi_2022_CVPR,khalid2022text,michel2022text2mesh,sanghi2022textcraft,jain2022zero,hong2022avatarclip,liu2022iss,gao2022get3d,poole2022dreamfusion,lin2022magic3d,alex2022pointe} show that shape generation and mesh stylization can be benefited from pre-trained large-scale language-image models such as CLIP~\cite{radford2021learning} or pre-trained text-to-image models, by leveraging rendered images of shapes as bridges. We notice that in many existing works, text or image inputs are converted to a single feature vector by the CLIP model; thus, it is hard to offer more local control on 3D synthesis. Our locally conditional mechanism is designed to remedy this issue for image conditioning.

\paragraph{Sketch-based shape reconstruction and generation} Many deep learning methods formulate sketch-to-3D task as shape reconstruction from single or multiple images~\cite{fan2016point,lun20173d,mescheder2019occupancy,xu2019disn,Li2018,saito2019pifu}, using 3D reconstruction losses~\cite{zhong2020deep} for training. With additional view information and 2D projection losses, some methods~\cite{liu2019soft,xiang2020sketch,wang2021multi,guillard2021sketch2mesh,zhang2021sketch2model,zhong2022study} show more promising reconstruction results. However, these deterministic approaches suffer from the ambiguity problem caused by single-view input. On the contrary, probabilistic generative methods can provide plausible outputs, as shown in \cite{autosdf2022,zhang20223dilg,chou2022diffusionsdf}, but they usually encode the input image as a global feature, and thus are difficult to provide local controllability and offer good generalizability to unseen shape variations. SketchSampler~\cite{gao2022sketchsampler} used the predicted density map as a proxy to improve reconstruction fidelity and combine noise sampling to predict depth values in a probabilistic generation way. However, it makes a strong assumption that input sketches are under orthogonal projection, and it is not easy to use their point cloud outputs for other applications.

\section{View-Aware Locally Attentional SDF Diffusion} \label{sec:method}

\subsection{Method Overview} \label{subsec:overview}

\begin{figure*}[t]
    \centering
    \includegraphics[width=\linewidth]{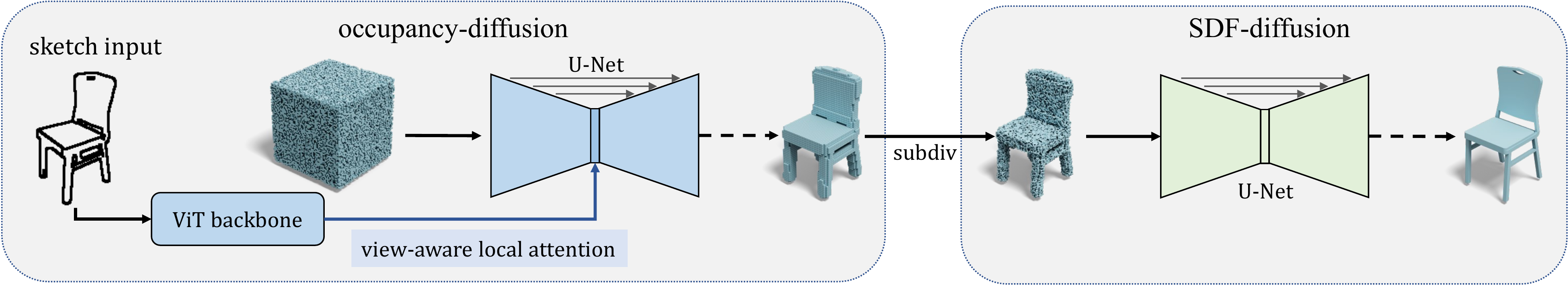}
    \caption{Our LAS-Diffusion model includes two stages: \emph{occupancy-diffusion} and \emph{SDF-diffusion}. \emph{Occupancy-diffusion} takes a noisy $64^3$ voxel grid as input, and uses a 3D U-Net to transform the volume to an occupancy volume. The occupied voxels are subdivided into a $128^3$ sparse voxel grid and filled with random noise. \emph{SDF-diffusion} takes this noisy sparse voxel grid as input, and transforms noise signals to SDF values via a 3D sparse-voxel-based U-Net. For sketch-conditional inputs, the local image patch features obtained from a pretrained ViT backbone interact with U-Net voxel features via a view-aware local attention mechanism, to offer local controllability and better generalizability. }
    \label{fig:overview} 
\end{figure*}

\paragraph{Discrete signed distance function} We choose discrete signed distance functions (SDF) as our 3D representation. A discrete signed distance function $g: \mathbf{z} \in \mathcal{Z} \mapsto \mathbb{R}$
is defined on a regular 3D grid $\mathcal{Z}$ or a subset of $\mathcal{Z}$. $g(\mathbf{z})$ records the signed distance from the centers of the grid cells to a closed manifold surface $\mathcal{S}$. Its zero isosurface in polygonal mesh format can be extracted from the dual grid of $\mathcal{Z}$ using the Marching Cube algorithm~\cite{lorensen1987marching}.

\paragraph{Discrete surface-occupancy function} A discrete signed distance function $g$ can be converted into a discrete surface occupancy function $o: \mathbf{z} \in \mathcal{Z} \mapsto  \{0, 1\}$ as follows:
$o(\mathbf{z}) = 1$ if $|g(\mathbf{z})| \leq \delta$; otherwise, $o(\mathbf{z}) = 0$. Here, $\delta > 0$ is the predefined threshold. The set of $\Omega_o := \{\mathbf{z} \in \mathcal{Z}: o(\mathbf{z}) = 1\}$ collects the grid cells whose shortest distance from their centers to the surface is no more than $\delta$. Here, note that $\Omega_o$ approximates the thin shell of a 3D shape only.

\paragraph{Two-stage diffusion}
To represent the details and small features of 3D shapes in discrete SDF format, a high-resolution grid is needed. However, it is not practical to generate a high-resolution and full-grid discrete SDF due to its cubic complexity in memory storage and computational cost.
To overcome this issue, we designed a two-stage generation framework based on a self-conditioning continuous diffusion model (\cref{subsec:diffusion}): The first stage generates a low-resolution discrete surface-occupancy function to approximate the shell of the shape (\cref{subsec:occdiffusion}), and the second stage focuses on generating fine-grained discrete SDF values inside the occupied region (\cref{subsec:sdfdiffusion}). We name these two stages by \emph{occupancy-diffusion} and \emph{SDF-diffusion}, respectively. In our implementation, the low resolution of the discrete surface-occupancy function is set to $64^3$, and the fine resolution of the discrete SDF is $128^3$.

\paragraph{Sketch-conditioned generation}
To incorporate 2D sketches as guidance, we use the local patch features of the sketch image to assist network learning in a view-aware and cross-attention manner (\cref{subsec:condition}). This mechanism is called \emph{view-aware local attention}.  Compared to using a global image feature as guidance, view-aware local attention provides better local controllability and makes the model generalizable to unseen sketches.

\cref{fig:overview} illustrates the pipeline of our two-stage diffusion. Each diffusion module is trained individually, and their network architectures are presented in the following subsections.

\subsection{Self-conditioning Continuous Diffusion Model} \label{subsec:diffusion}

\paragraph{Continuous denoising diffusion}
A typical continuous denoising diffusion model~\cite{sohl2015deep,ho2020denoising,kingma2021variational} consists of the forward process and the reverse process. The forward process introduces a sequence of increasing (Gaussian) noise to a data point $\boldsymbol{x}_0$, such that it ends up at $\boldsymbol{x}_t$ that follows the predefined Gaussian distribution. Here, $t$ runs from 0 to 1 in a continuous way. The reverse process maps a noise $\boldsymbol{\epsilon}$ sampled from a Gaussian distribution to a data point $\boldsymbol{x}_0$ through a series of state transitions.
The forward process from $\boldsymbol{x}_0$ to $\boldsymbol{x}_t$ can be defined as follows.
\begin{equation}
  \boldsymbol{x}_t=\sqrt{\gamma(t)} \boldsymbol{x}_0+\sqrt{1-\gamma(t)} \boldsymbol{\epsilon},
  \label{eq:diffusion_forward}
\end{equation}
where $\boldsymbol{\epsilon} \sim \mathcal{N}\left(\boldsymbol{0}, \boldsymbol{I}\right), t \sim \mathcal{U}(0, 1)$, and $\gamma(t)$ is a monotonically decreasing function from 1 to 0. $\mathcal{N}$ and $\mathcal{U}$ denote Gaussian distribution and uniform distribution, respectively.
In our implementation, we follow \cite{kingma2021variational} to set $\gamma(t) = e^{-10t^2-10^{-4}}$.

The prediction from $\boldsymbol{x}_t$ to $\boldsymbol{x}_0$ can be modeled by a neural network $f(\boldsymbol{x}_t, t)$. The network training is based on the following denoising loss:
\begin{equation}
  \mathcal{L}_{\boldsymbol{x}_0}=\mathbb{E}_{\boldsymbol{\epsilon} \sim \mathcal{N}\left(\boldsymbol{0}, \boldsymbol{I}\right),t \sim \mathcal{U}\left(0, 1\right)}\left\|f\left(\boldsymbol{x}_t, t\right)-\boldsymbol{x}_0\right\|_2^2.
  \label{eq:diffusion_loss}
\end{equation}
Here, $\boldsymbol{x}_t$ is sampled via \cref{eq:diffusion_forward}.
$f$ is usually implemented as a U-Net architecture. The sampling methods such as DDPM~\cite{ho2020denoising} and DDIM~\cite{song2020denoising} strategy can be used for sample generation. For conditioned 3D generation, we adopt the classifier-free guidance~\cite{ho2021classifierfree} technique.

\paragraph{Self-conditioning} Recently, Chen \etal~\shortcite{chen2022analog} introduced the self-conditioning mechanism, which uses the previously generated samples as conditioning to significantly improve diffusion models. This mechanism is simple: a neural network $f(\boldsymbol{x}_t, \tilde{\boldsymbol{x}}_0, t)$ is trained to map $\boldsymbol{x}_t$ to $\boldsymbol{x}_0$, where $\tilde{\boldsymbol{x}}_0$ is an estimated $\boldsymbol{x}_0$ from the previous prediction.  The loss function \cref{eq:diffusion_loss} is revised as follows.
\begin{equation}
  \mathcal{L}_{\boldsymbol{x}_0}=\mathbb{E}_{\boldsymbol{\epsilon} \sim \mathcal{N}\left(\boldsymbol{0}, \boldsymbol{I}\right),t \sim \mathcal{U}\left(0, 1\right)}\left\|f\left(\boldsymbol{x}_t, \tilde{\boldsymbol{x}}_0, t\right)-\boldsymbol{x}_0\right\|_2^2.
  \label{eq:diffusion_loss_revised}
\end{equation}
As suggested by \cite{chen2022analog}, during network training, $\tilde{\boldsymbol{x}}_0$ is set to $f\left(\boldsymbol{x}_t,\boldsymbol{0}, t\right)$ with probability $p$, and  $\boldsymbol{0}$ with probability $1 - p$, \ie without self-conditioning. Here, $p$ is set to $0.5$ by default. Gradient backpropagation on $\tilde{\boldsymbol{x}}_0$ is disabled to reduce the total training time.

\subsection{Occupancy-diffusion Module} \label{subsec:occdiffusion}
Our occupancy-diffusion module is designed to transfer a noisy coarse grid to a discrete surface-occupancy function of a 3D shape.  In the following, we introduce the creation of ground-truth discrete surface-occupancy functions and the network architecture.

\paragraph{Data preparation}
For each 3D shape in the dataset, we normalize it to fit in a $[-0.8,0.8]^3$ box, and compute the discrete SDF function with resolution $128^3$ in $[-1,1]^3$ using the algorithm of \cite{xu2014signed}. This step is similar to \cite{zheng2022sdfstylegan}.
Based on the fact that any voxel in a $64^3$ grid contains 8 subvoxels of the $128^3$ grid. We create a discrete surface-occupancy function $o$ in $64^3$ resolution as follows:  $o(\mathbf{z}) = 1$ if there exists a subvoxel of $\mathbf{z}$ whose stored SDF value $v$ satisfies $|v| \leq \frac{1}{32}$; otherwise, $o(\mathbf{z})$ is set to 0.

\paragraph{Network architecture}
The U-Net structure in the module is built on the standard 3D convolutional neural network. The U-Net has 5 levels: $64^3$, $32^3$, $16^3$, $8^3$, $4^3$, and the feature dimensions are $32$, $64$, $128$, $256$, and $256$, respectively. Each level is made up of a ResNet block that contains two convolution layers with kernel size 3. In the bottleneck of U-Net, we add two ResNet blocks. A convolution layer is attached at the end of the network to map the voxel features at the finest level to a surface-occupancy value.

\paragraph{Network training} \cref{eq:diffusion_loss_revised} is used for network training. More specifically, $\boldsymbol{x}_0$ is the tensor that stores the ground-truth discrete surface-occupancy values of the grid.  As we use the self-conditioning continuous diffusion model, the estimated $\tilde{\boldsymbol{x}}_0$ is treated as an additional input channel to the U-Net.

\paragraph{Network inference} The $64^3$ grid is first initialized with Gaussian noise, then we denoise it in a finite number of steps, using the DDPM sampling strategy~\cite{ho2020denoising}. We reserve the voxels whose predicted surface-occupancy values are larger than $0.5$, and subdivide them once to obtain a set of subvoxels in $128^3$ resolution.


\subsection{SDF-diffusion Module} \label{subsec:sdfdiffusion}
For a set of sparse voxels with noisy SDF values, our SDF-diffusion module is designed to map it to the discrete SDF function that represents a real shape. We use the $128^3$ discrete SDF functions as described in \cref{subsec:occdiffusion} for training. The U-Net structure is similar to the one used for occupancy-diffusion except that (1) we use octree-based convolution neural network~\cite{Wang2017,wang2020} as the SDF data is stored in sparse voxel format; (2) the U-Net has 4 levels: $128^3$, $64^3$, $32^3$, $16^3$, and the feature dimensions are $32,64,128,256$, respectively. The network training is similar to occupancy-diffusion, and $\boldsymbol{x}_0$ in the loss function corresponds to
the SDF values stored at the finest octree nodes.

\paragraph{Network inference} The subdivided voxels from occupancy-diffusion are initialized with Gaussian noise, we denoise it through the DDPM sampling strategy and apply the Marching Cube algorithm~\cite{lorensen1987marching} on the dual grids of the resulting discrete SDF function to obtain the mesh output.

\subsection{View-aware Local Attention} \label{subsec:condition}
Our approach supports sketch-conditioned shape generation by a novel view-aware local attention mechanism. For a sketch image input, we assume that the view information of the sketch is known, \ie camera position and orientation with respect to the shape in a canonical pose. Thus, we can align the image and the 3D grid volume according to the view projection.  Inspired by the works of \cite{wang2018pixel2mesh,xu2019disn} that leverage pixel-level image features for shape reconstruction, we propose to use local image patch features to guide surface-occupancy generation, via feature cross-attention, as follows.

\paragraph{Patch feature extractor} We choose the vision-transformer (ViT) backbone pre-trained on a large volume of images as our sketch image feature extractor. The ViT backbone represents an input image as a series of non-overlapped image patches, denoted by $P_1, P_2, \cdots$, and encodes the image into a set of patch-wise features~\cite{dosovitskiy2020image}.

\begin{figure}[t]
    \centering
    \begin{overpic}[width=0.85\linewidth]{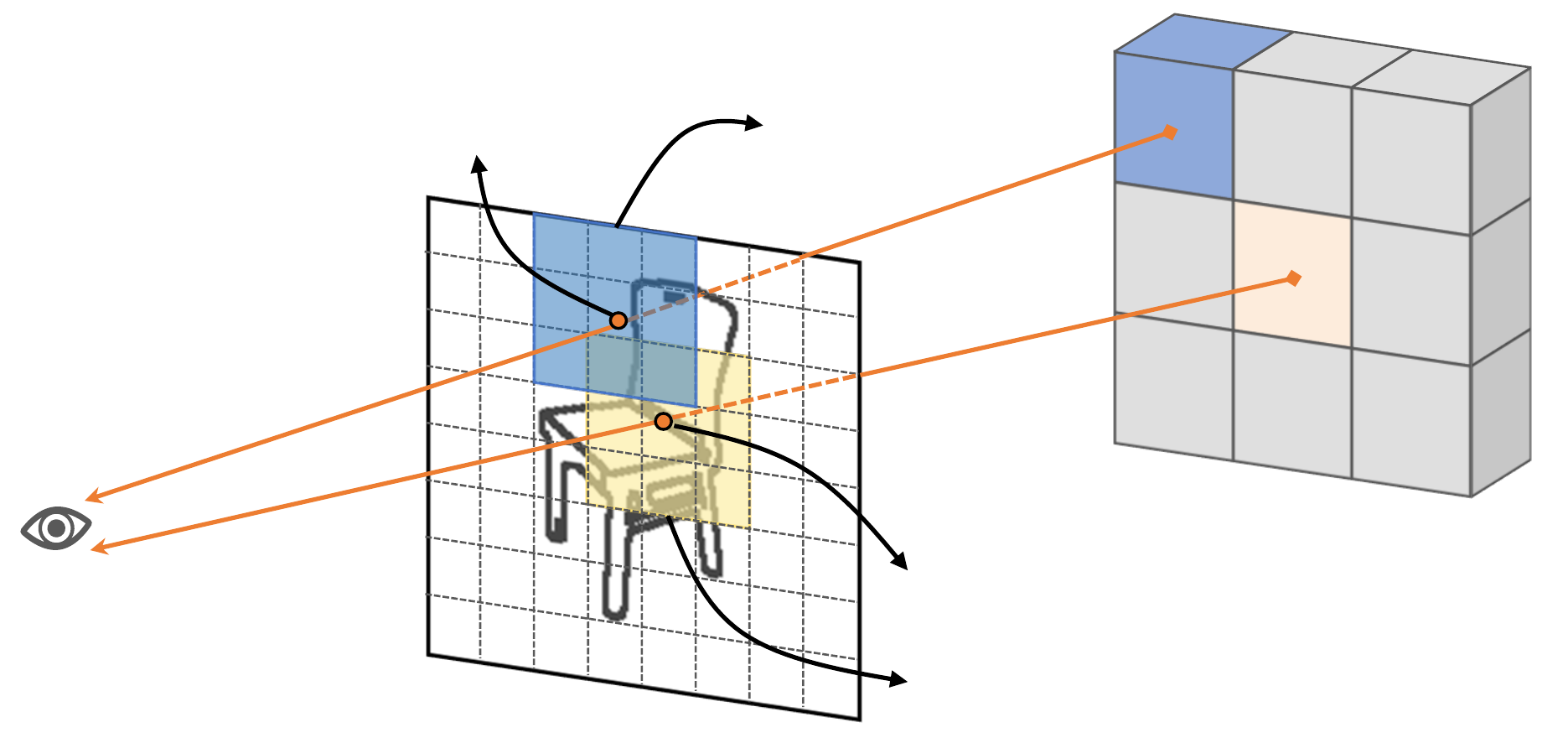}
        \put(80,30){\small $V$}
        \put(72,40.5){\small $U$}
        \put(58,8.5){\small $\mathbf{p}$}
        \put(58,2.5){\small $\mathcal{N}_{V}$}
        \put(29,40){\small $\mathbf{q}$}
        \put(49,39){\small $\mathcal{N}_{U}$}
        \put(31, -2){\small image plane}
        \put(75, 11.5){\small voxel grid}
    \end{overpic}
    \caption{Illustration of our view-aware local attention mechanism. For voxel $V$, its voxel center is projected onto the image plane at $\mp$, via a known perspective projection. We use the image patch features of the local patches around $\mp$ (in yellow color), to interact with voxel feature at $V$ in the U-Net, via cross-attention. For other voxels such as $U$, the operation is similar.}
    \label{fig:cross_attention}
\end{figure}
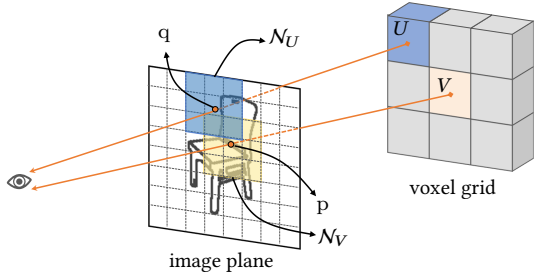 

\paragraph{View-aware local attention} We let the voxel features in the U-Net of the occupancy-diffusion module interact with the image patch features according to their view-projection-based relationship. For any voxel $V$ in the grid, we project its center onto the sketch image and obtain the projected coordinate $\mathbf{p}$. Neighboring image patches close to $\mathbf{p}$ are selected to interact with $V$ because their features are highly likely to affect local geometry controlled by $V$. The set of neighborhood image patches is denoted by $\mathcal{N}_V$, and selected as follows: \emph{patch $P_j$ belongs to $\mathcal{N}_V$ if the distance between $\mathbf{p}$ and the center of $P_j$ is less than a distance threshold $d_\delta$}.
\cref{fig:cross_attention} illustrates the relationship between a voxel and its related image patches.

We use one-layer multi-head cross-attention~\cite{vaswani2017attention} to model feature interaction between the voxel feature $f_V$ at $V$ and the set of image patch features $f_{I}$ that belongs to the patches in $\mathcal{N}_V$, as follows.
\begin{equation}
  \begin{aligned}
    Q = f_V\mathbf{W}^Q, K = f_\mathcal{N}\mathbf{W}^K, V = f_\mathcal{N}\mathbf{W}^V; & \\
    f_V^{\text{new}} = \text{MH-Attention}\left(Q,K,V,\mathcal{M}\right).              &
  \end{aligned}
  \label{eq:crossattn}
\end{equation}
Here, $\text{MH-Attention}(\cdot)$ is the standard multi-head attention operation, $\mathcal{M}$ is the mask for attention calculation induced by view projection, and we use absolute positional encoding for both voxels and image patches.

Due to the use of patch features, our view-aware local attention is not sensitive to small errors of projection views, as a small view perturbation may still result in similar local patch sets. Therefore, our design is friendly to sketch-conditioned 3D generation, and the user only needs to provide a rough guess of view information, either by a manual way or with the aid of a view prediction network.

\paragraph{Implementation} We use a huge ViT model pre-trained on Laion2B dataset~\cite{laion2B} as our ViT backbone.
Its default input image resolution is $224\times 224$ and the \texttt{patch\_width} is $14$.  The default value of $d_\delta$ is set to $4 \times \texttt{patch\_width}$.
The weights of the ViT backbone are frozen for our use. To reduce computational cost, only the voxels at $8^3$ and $4^3$ levels are involved in view-aware local attention. We also experienced using view-aware local attention in our SDF-diffusion module, but found that it has limited improvement to local geometry; thus we introduce view-aware local attention to occupancy-diffusion only.

\def \viewzero {\texttt{left}}
\def \viewone {\texttt{side-left}}
\def \viewtwo {\texttt{front}}
\def \viewthree {\texttt{side-right}}
\def \viewfour {\texttt{right}}

\section{Sketch-conditioned Shape Generation} \label{sec:sketchcondition}
In this section, we exhibit and validate the capability of LAS-Diffusion for sketch-conditioned shape generation.

\subsection{Model Training}

\begin{figure}[t]
    \centering
    \includegraphics[width=0.9\linewidth]{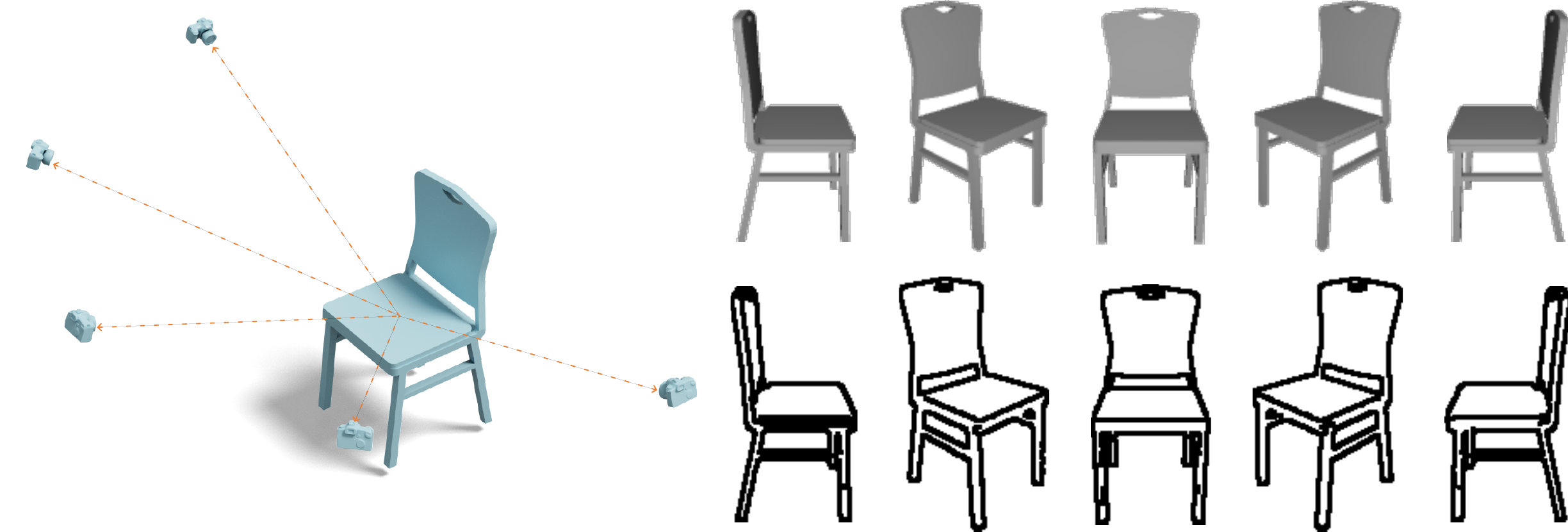}
    \caption{\textbf{Left}: Camera setup. \textbf{Right}: Shading images and sketches under the predefined views.
    }
    \label{fig:viewselection} 
\end{figure}

\paragraph{Training dataset} We choose 5 categories from ShapeNetV1~\cite{chang2015shapenet}: \texttt{chair}, \texttt{car}, \texttt{airplane}, \texttt{table}, and \texttt{rifle} for training our sketch-conditioned model.

\paragraph{Predefined perspective views} To make our model convenient to use, we provide five perspective views for user selection and prepare the corresponding sketch data for training. For a normalized shape in its canonical pose, we place five cameras on the scaled bounding sphere of the shape, as shown in \cref{fig:viewselection}-left.
These five perspective views are chosen because the sketches under these views are informative and normal users tend to use these view directions or their nearby view directions to draw sketches. The user can pick one of the predefined views that best matches the input sketches for model inference. For convenience, we denote these views by \viewzero, \viewone, \viewtwo,  \viewthree, and \viewfour. \cref{fig:viewselection} illustrates shading images and sketches of a chair model under these predefined views.

\paragraph{Data preparation}
For each shape in the dataset, we render its shading images from different views and extract their edges via Canny edge detector~\cite{canny1986computational}, as 2D sketches. Note that other kinds of sketch synthesis techniques can be used, and we use Canny edge for simplicity. We restrict the rendered views to the predefined views with random perturbation to improve the robustness of the network. The random perturbation is implemented by perturbing the azimuth angle by a noise within $\pm \ang{22.5}$ and the elevation angle by a noise within $\pm\ang{5}$. For each predefined view, we perturb $10$ times. In total, there are $50$ sketches for a shape. During training, the corresponding views of these sketches are grouped into their predefined perspective views. To enhance local prior learning, we also augmented shape data by simply uniting two randomly selected shapes for occupancy-diffusion module training, where the shapes are also translated randomly. The number of augmented shapes is the same as the number of original shapes. We trained a single sketch-conditioned LAS-Diffusion model on the five chosen shape categories.

\paragraph{Training details}
We trained the occupancy-diffusion module using Adam optimizer~\cite{kingma2014adam} with a fixed learning rate of $2\times 10^{-4}$ over $300$ epochs. For the training of the SDF-diffusion module, we used AdamW optimizer~\cite{loshchilov2018decoupled} with a fixed learning rate of $10^{-4}$ over $500$ epochs, and its training split follows \cite{chen2019learning}.

\paragraph{Inference efficiency} The inference time of our model on a machine with an Nvidia 1080 Ti GPU takes around 10 seconds, using a 50-step DDPM sampling strategy.

\paragraph{Competing methods} We choose the following representative sketch-to-3D methods for comparison: Sketch2Model~\cite{zhang2021sketch2model}, Sketch2Mesh~\cite{guillard2021sketch2mesh} and SketchSampler~\cite{gao2022sketchsampler}. Sketch2Model has trained its category-specific models on 13 ShapeNet categories and Sketch2Mesh has trained its category-specific models on \texttt{car} and \texttt{chair} categories. SketchSampler has trained a single model on 13 ShapeNet categories. As SketchSampler is only capable of producing point clouds, we convert them to meshes using SAP~\cite{peng2021shape} for quantitative evaluation. For all the above methods, we use their pre-trained models for comparison.

\paragraph{Evaluation metrics}
Since there is no previous work designing evaluation metrics for sketch-conditioned probabilistic generative methods, we adapt \emph{CLIP score}~\cite{hessel2021clipscore} to evaluate perception difference as follows.
For a generated 3D shape conditioned on a sketch $I$, we render its sketch $G$ under the same view of the input sketch by our data preparation pipeline, then compute the cosine similarity between the clip features of these two sketches:
\begin{equation}
    \text{CLIPScore}(I,G) = 100 \times \langle E_I, E_G \rangle.
\end{equation}
Here, $E_I$ and $E_G$ are the normalized CLIP features of $I$ and $G$, respectively, and $\langle\cdot,\cdot\rangle$ is inner product. The score is averaged over all test data.
We also treat the non-white pixels of sketches as 2D points and measure the 2D Chamfer distance between $I$ and $G$, denoted by \emph{Sketch-CD}. The reconstruction metrics such as CD, EMD, and Voxel-IOU from SketchSampler~\cite{gao2022sketchsampler} are also adopted to evaluate the 3D quality of the generated or reconstructed shapes with respect to the shapes that the sketches correspond to.
\subsection{Model Evaluation}\label{sec:model_evalution}
\begin{table}[t]
    \caption{Quantitative evaluations on IKEA chairs. The units of Sketch-CD, CD, EMD and Voxel-IOU are  $10^{-4}$, $10^{-3}$, $10^{-2}$ and $10^{-2}$, respectively. SketchSampler$_m$ denotes the version that the point cloud outputs of SketchSampler is converted to polygonal meshes.}
    \label{tab:ikea_sketch}
    \centering
    \resizebox{\columnwidth}{!}
    {
        \begin{tabular}{c|cc|ccc}
            \toprule
            \thead{Method}    & \thead{CLIPScore$\uparrow$} & \thead{Sketch-CD$\downarrow$} & \thead{CD$\downarrow$} & \thead{EMD$\downarrow$} & \thead{Voxel-IOU$\uparrow$} \\ \midrule
            Sketch2Model      & 88.77                       & 101.0                         & 49.38                  & 20.31                   & 22.76                       \\
            Sketch2Mesh       & 93.46                       & 37.64                         & 19.16                  & 16.39                   & 32.40                       \\
            SketchSampler     & N/A                         & N/A                           & 32.51                  & 20.24                   & 33.82                       \\
            SketchSampler$_m$ & 90.43                       & 42.94                         & 33.41                  & 21.24                   & 26.67                       \\
            LAS-Diffusion     & \textbf{96.92}              & \textbf{10.33}                & \textbf{{\color{white}0}6.48}      & \textbf{{\color{white}0}8.85}           & \textbf{49.83}              \\
            \bottomrule
        \end{tabular}
    }
\end{table}

\begin{figure}[t]
    \centering
    \begin{overpic}[width=0.95\linewidth]{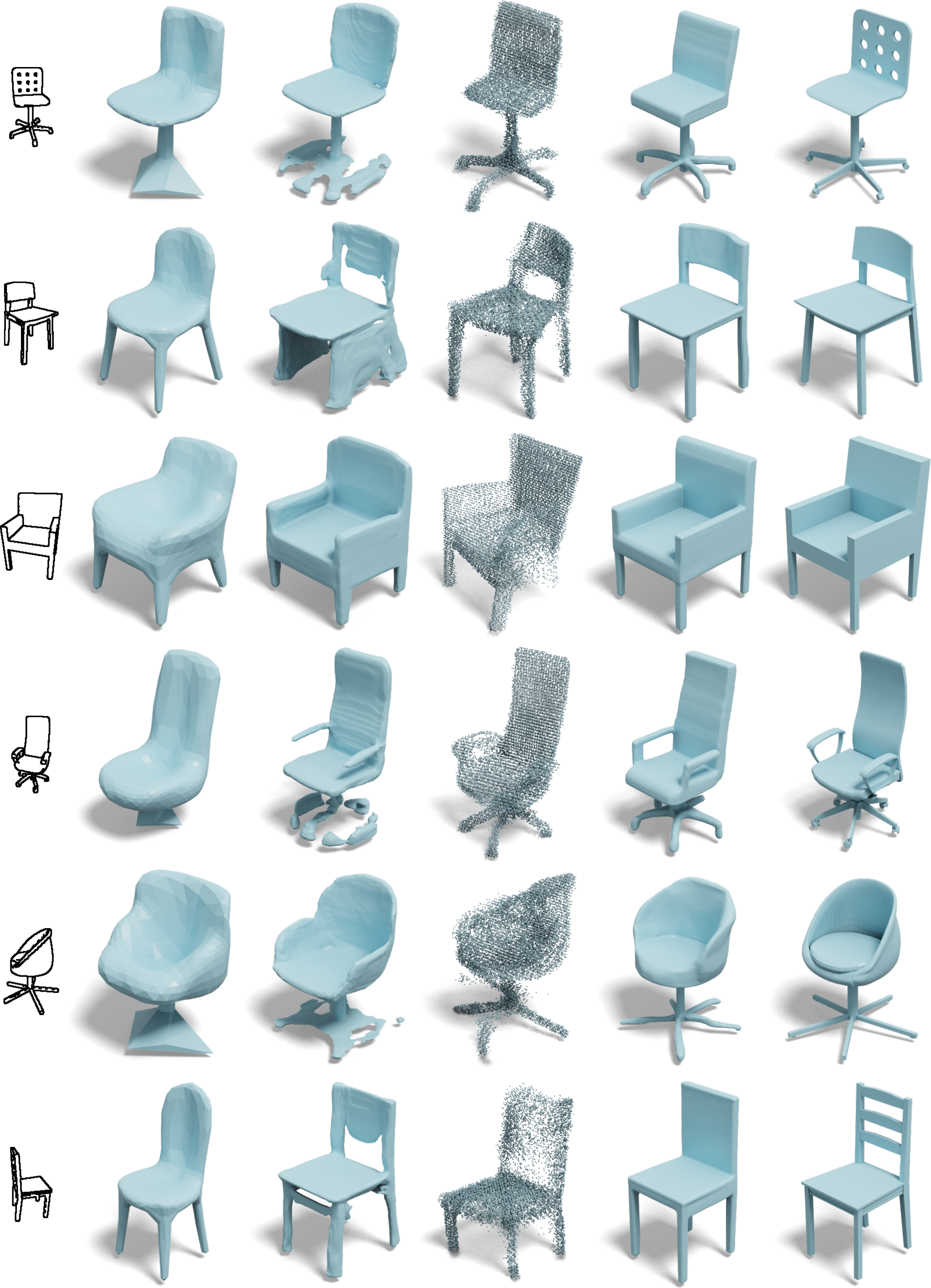}
        \put(7, -3){\scriptsize \textbf{Sketch2Model}}
        \put(20.8, -3){\scriptsize \textbf{Sketch2Mesh}}
        \put(34, -3){\scriptsize \textbf{SketchSampler}}
        \put(48, -3){\scriptsize \textbf{LAS-Diffusion}}
        \put(66, -3){\scriptsize \textbf{GT}}
    \end{overpic}
    \caption{Sketch-conditioned shape generation on IKEA chairs.}
    \label{fig:ikea}
\end{figure}

\paragraph{Quantitative and qualitative evaluation} We choose IKEA~\cite{IKEA} chair dataset as the test bed, which contains $35$ chairs. Each chair is rendered from a random view to generate a sketch. For our model, we use one of the predefined views that best matches the input sketch for model inference.  We provide accurate view information to Sketch2Model and Sketch2Mesh, and use suggestive~\cite{decarlo2003suggestive} algorithm to prepare sketches for Sketch2Mesh to match its training style. Compared with other methods, our LAS-Diffusion achieves significantly better performance, as reported in \cref{tab:ikea_sketch}. \cref{fig:ikea} visualizes the results of different methods. The results of LAS-Diffusion are the most plausible and possess better geometry quality. We also render the synthetic sketches of their results in \cref{fig:sketch_compare}, and find that our results match better with the input sketch. In the supplemental material, we provide all of our generation results.

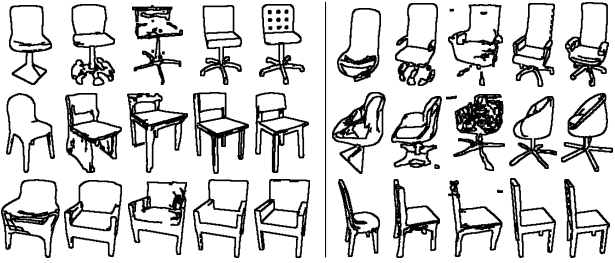
\begin{figure}[t]
    \centering
    \begin{overpic}[width=0.95\linewidth]{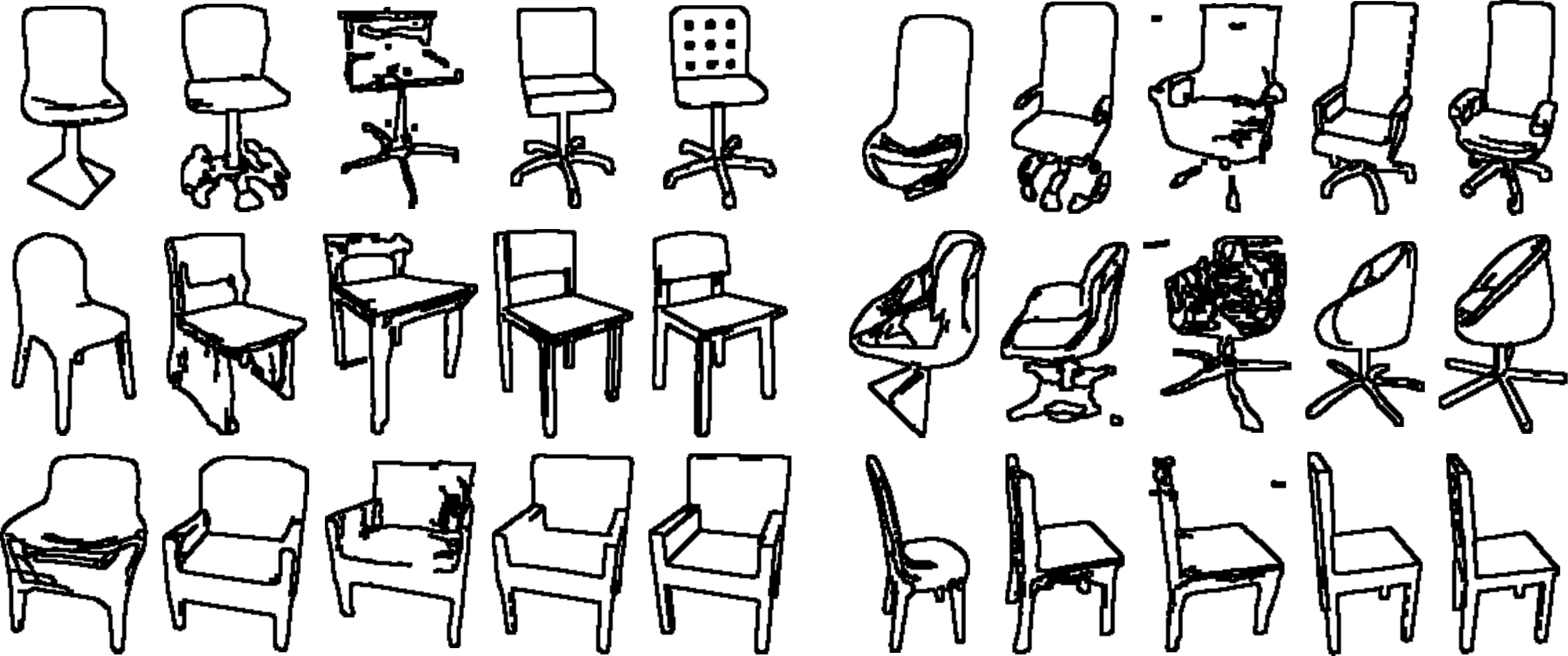}
        \put(52,0){
            \begin{tikzpicture}
                \draw (0,0) -- (0,3.4);
            \end{tikzpicture}
        }
    \end{overpic}
    \caption{The sketches of the results shown in \cref{fig:ikea}. From \textbf{left} to \textbf{right}: Sketch2Model, Sketch2Mesh, SketchSampler, LAS-Diffusion and GT.}
    \label{fig:sketch_compare} 
\end{figure} 
\begin{figure}[t]
    \centering
    \includegraphics[width=0.95\linewidth]{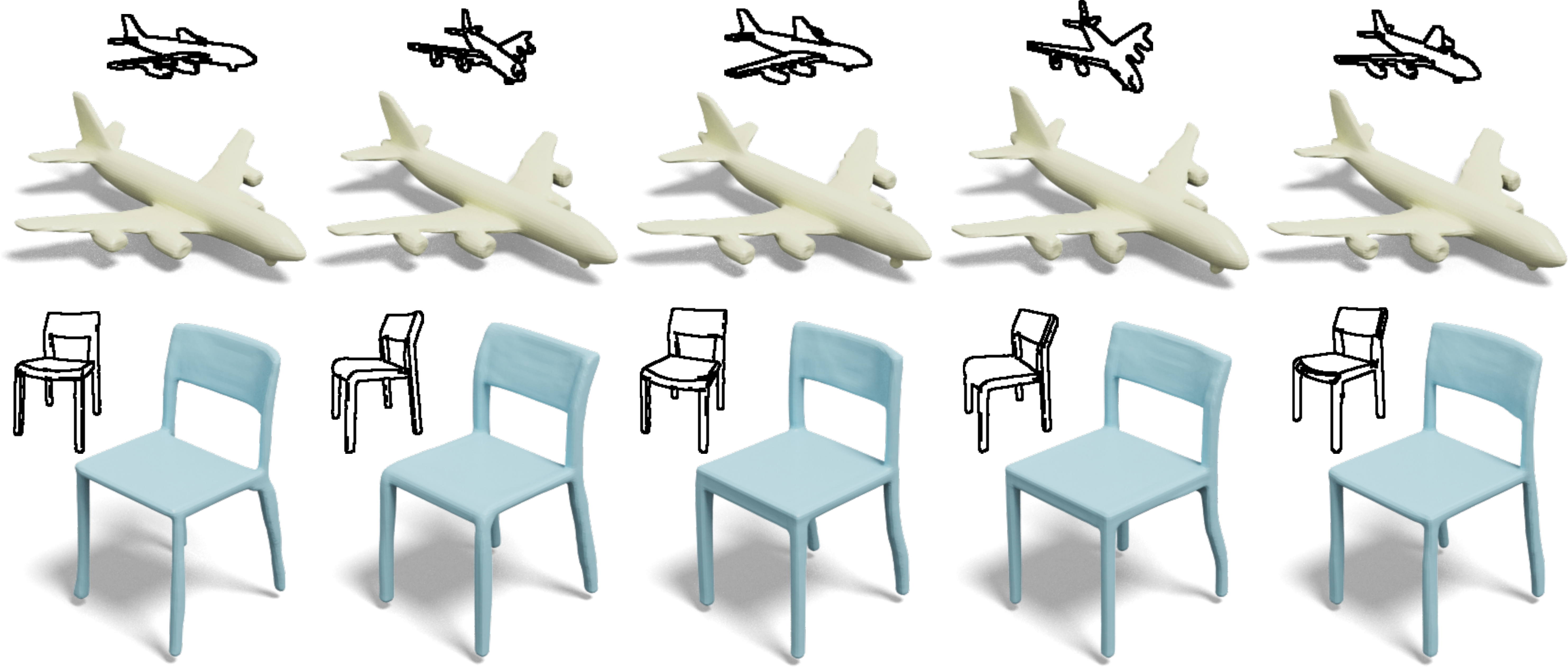}
    \caption{View robustness test. The views for model inference are \viewone{}  (top row) and \viewthree{} (bottom row).
    }
    \label{fig:view_robustness} 
\end{figure}
\begin{figure}[t]
    \centering
    \includegraphics[width=0.95\linewidth]{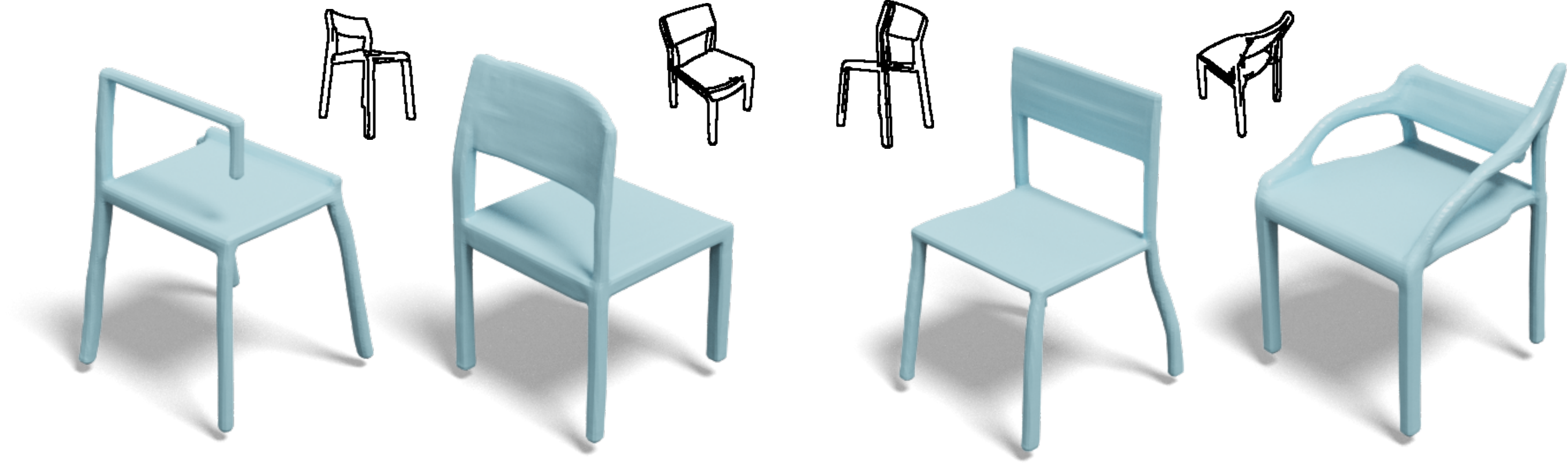}
    \caption{Stress test on view robustness. The view for model inference is \viewthree.}
    \label{fig:view_rob_failure} 
\end{figure}

\paragraph{View robustness} Although only predefined views can be used in our model inference, our model is robust to small view perturbation due to the use of image patch features and random view perturbation during training. In \cref{fig:view_robustness}, we illustrate model robustness on two test cases. For each case, we provide 5 different input sketches and we use their most similar view for model inference. We can see that the generated shapes have consistent and good geometry.  We also add a stress test in which the input sketches are very different from the predefined \viewthree{} view (see \cref{fig:view_rob_failure}), our model is still capable of producing chair-like shapes although some parts are distorted and incomplete due to the use of wrong view information.

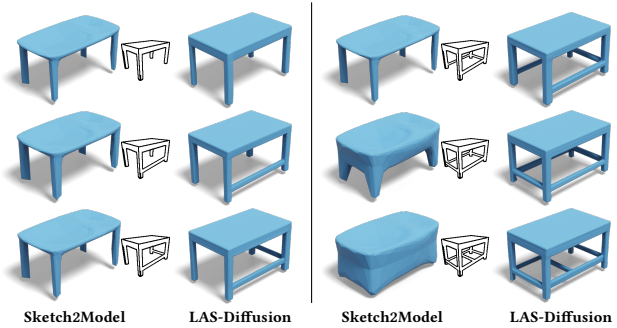
\begin{figure}[t]
    \centering
    \begin{overpic}[width=0.95\linewidth]{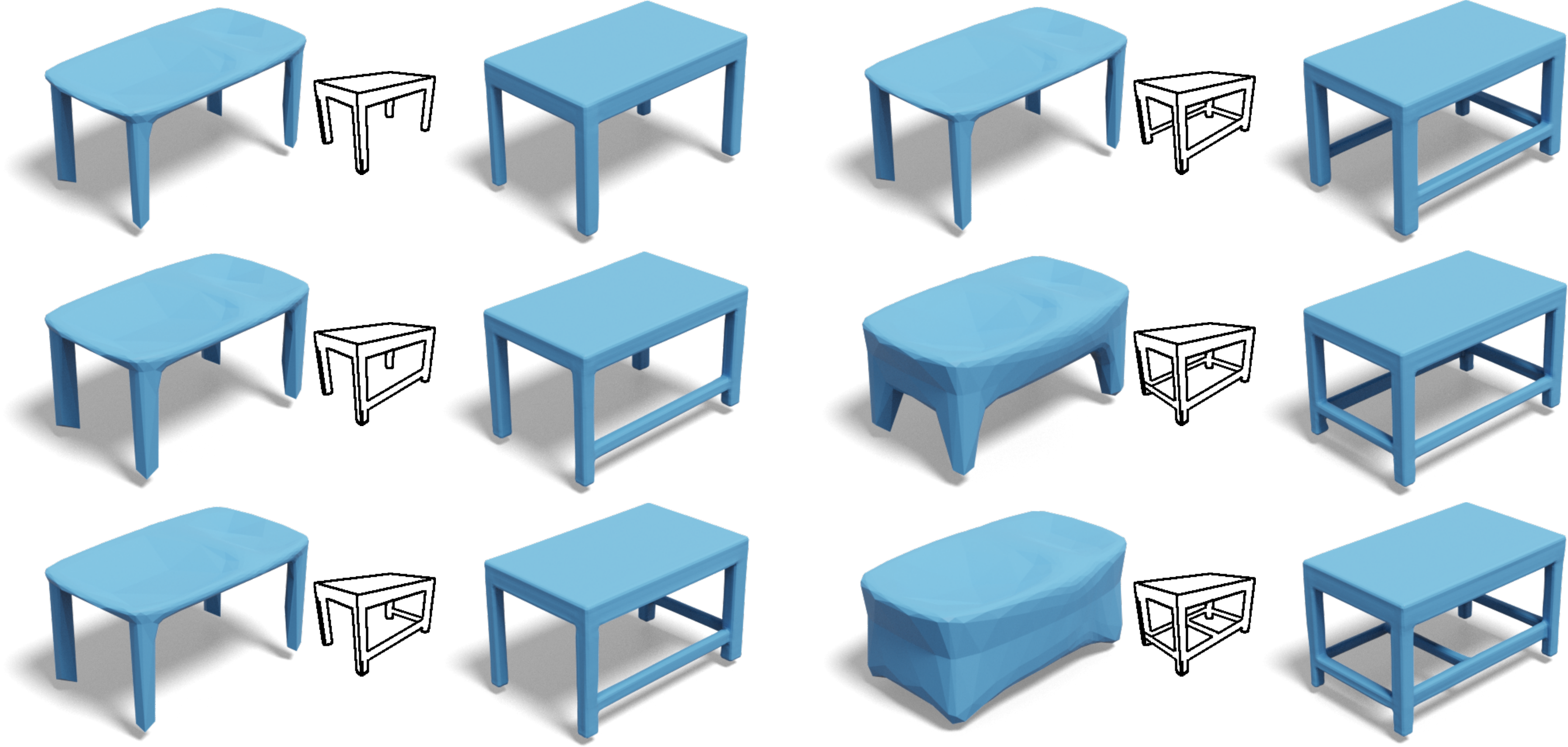}
        \put(4,-3){\scriptsize \textbf{Sketch2Model}}
        \put(31,-3){\scriptsize \textbf{LAS-Diffusion}}
        \put(56,-3){\scriptsize \textbf{Sketch2Model}}
        \put(83.5,-3){\scriptsize \textbf{LAS-Diffusion}}
        \put(50,0){
            \begin{tikzpicture}
                \draw (0,0) -- (0,4);
            \end{tikzpicture}
        }
    \end{overpic}
    \caption{The six input sketches have different numbers of horizontal bars. LAS-Diffusion has a high probability to generate 3D shapes that match the sketch inputs.}
    \label{fig:add-bar}
\end{figure} 

\paragraph{Local controllability} The view-aware local attention mechanism of our LAS-Diffusion model offers nice local controllability, as demonstrated by the example shown in \cref{fig:add-bar}, where a table sketch is modified to have different numbers of horizontal bars. LAS-Diffusion captures local changes well and has a high probability of generating structurally correct and geometry-plausible results.  In contrast, the reconstruction-based approach --- Sketch2Model~\cite{zhang2021sketch2model} cannot handle these structural changes well.

\begin{figure}[t]
    \centering
    \includegraphics[width=0.95\linewidth]{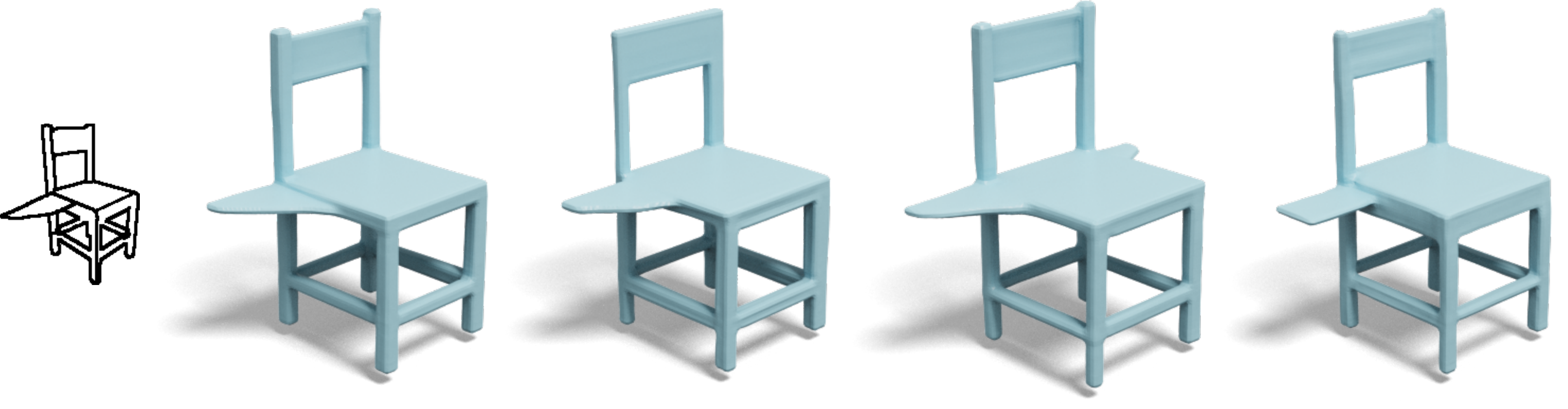}
    \vspace{-2mm}
    \caption{Shape generation conditioned on a creative sketch input.}
    \label{fig:generation_diversity} 
\end{figure} 
\paragraph{Model generalizability} We evaluate our model generalizability from the following four aspects.

\begin{figure}[t]
    \centering
    \begin{overpic}[width=0.95\linewidth]{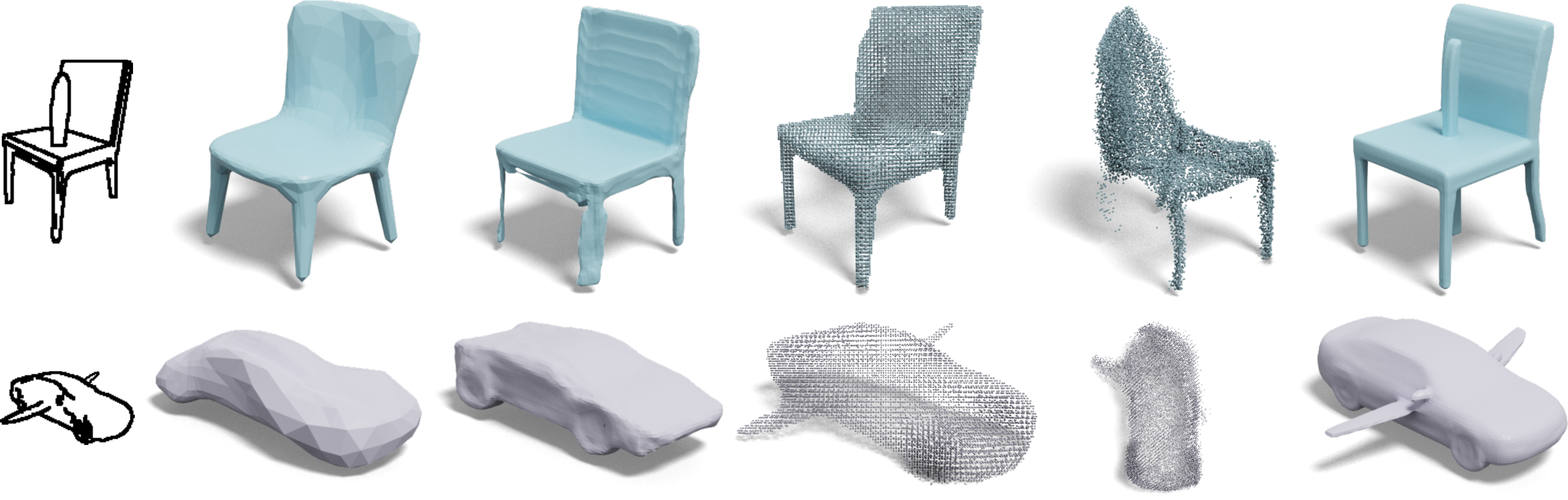}
        \put(10,-3){\scriptsize \textbf{Sketch2Model}}
        \put(30,-3){\scriptsize \textbf{Sketch2Mesh}}
        \put(56,-3){\scriptsize \textbf{SketchSampler}}
        \put(85,-3){\scriptsize \textbf{LAS-Diffusion}}
    \end{overpic}
    \caption{Comparisons of model generalizability. The results of SketchSampler~\cite{gao2022sketchsampler} are rendered from two different views, for better visualization.
    }
    \label{fig:generality-compare} 
\end{figure}

\emph{1. Unseen structural variations}. As LAS-Diffusion utilizes local image priors, it is well suited to generate 3D shapes with unseen structural variations. \cref{fig:generation_diversity} demonstrates this model generalizability by using a creative sketch input where a chair is attached with a wing-like part. We generate four results using LAS-Diffusion with different noises. The wing part appears in all the results, with some geometry variations.  \cref{fig:generality-compare} shows two more examples conditioned by creatively designed sketches. We also tested Sketch2Model, Sketch2Mesh, and SketchSampler on them. Both Sketch2Model and Sketch2Mesh fail to reconstruct the geometry unseen by their training set. SketchSampler has better generalizability due to the use of view-dependent depth sampling, but its predicted depth values do not meet the expectation and the quality of its output point clouds is low.

\begin{figure}[t]
    \centering
    \includegraphics[width=0.95\linewidth]{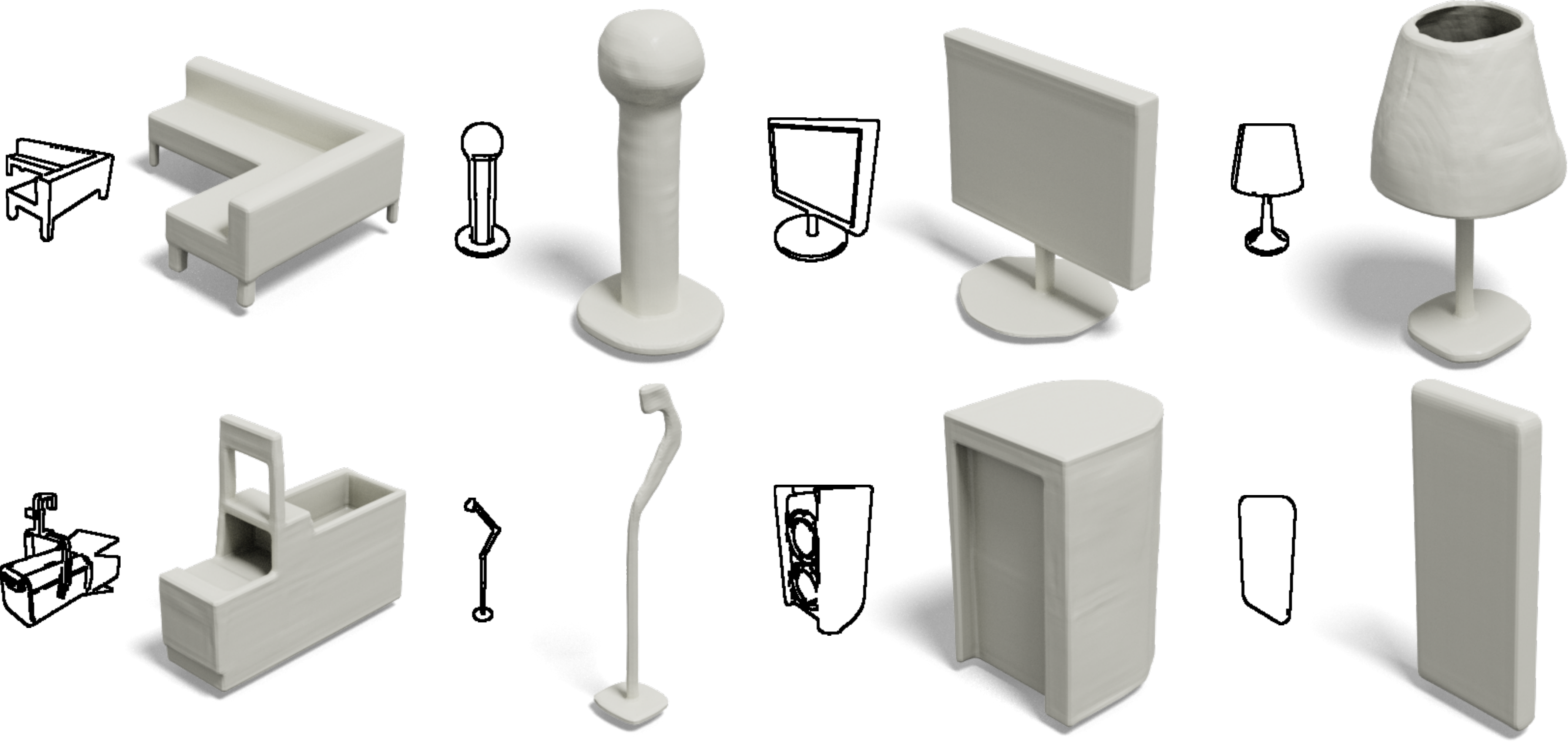}
    \caption{Our LAS-Diffusion model exhibits good generalizability to the sketches beyond the training categories.}
    \label{fig:outoddomain} 
\end{figure}

\emph{2. Unseen shape categories}. In \cref{fig:outoddomain}, we tested sketches of some objects that do not belong to the categories we trained. LAS-Diffusion can generate meaningful results. We attribute its success to local prior learning enabled by our view-aware local attention mechanism, and speculate that the local geometry shown in these examples may exist in the training data.

\begin{figure}[t]
    \centering
    \includegraphics[width=0.95\columnwidth]{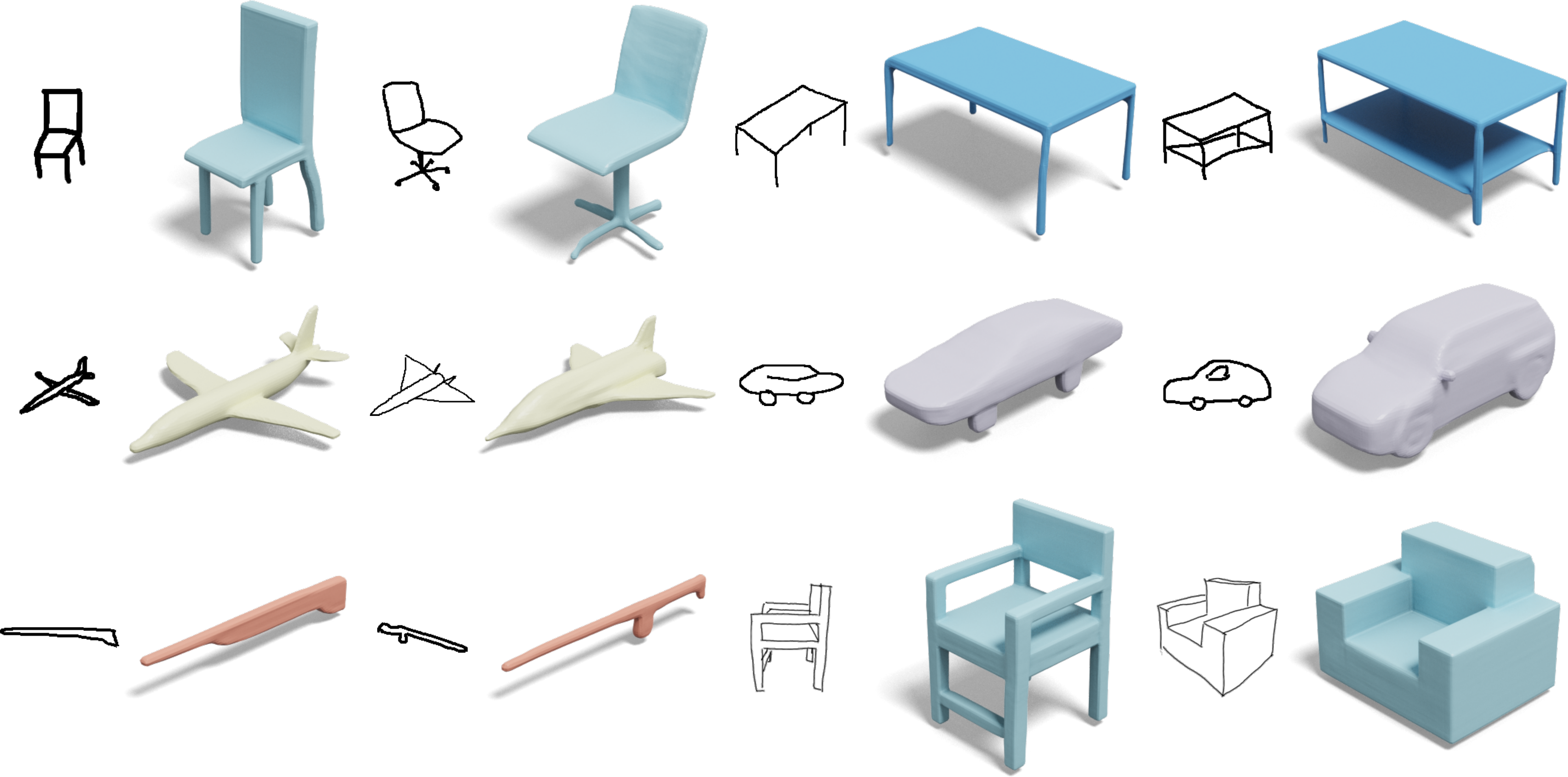}
    \caption{Our LAS-Diffusion model is capable of supporting freehand sketches.}
    \label{fig:humandrawing}  
\end{figure}

\emph{3. Freehand sketches}. Due to our local attention mechanism and the use of the pre-trained ViT image encoder, our model is tolerant of imprecise sketches and varied stroke widths that are different from our rendering setting, thus supporting freehand sketch input. In \cref{fig:humandrawing}, we provide some freehand-sketch-conditioned results.

\begin{figure}[t]
    \centering
    \includegraphics[width=0.95\linewidth]{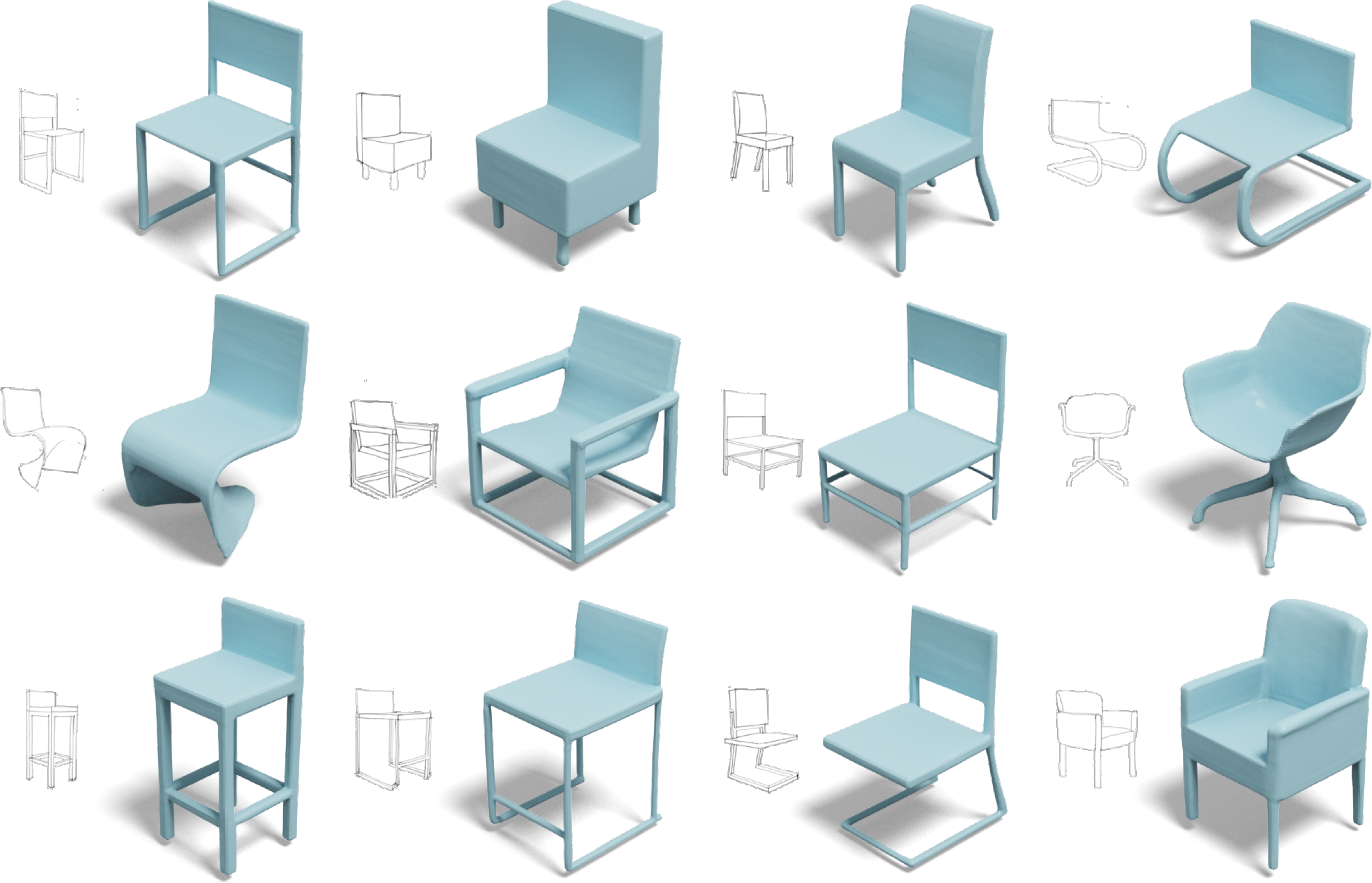}
    \caption{Our model supports sketches drawn by professional artists and generates plausible results. The sketches are from the ProSketch-3D dataset. }
    \label{fig:prosketch} 
\end{figure}

\emph{4. Professional sketches}. Sketch styles from professional artists are different from our synthetic sketches. We tested the robustness of our model to professional sketches using the ProSketch-3D dataset~\cite{zhong2020towards} which contains $500$ chairs and $1500$ sketches drawn by artists. As the elevation angle of the sketch view of ProSketch-3D data has a $\ang{20}$ difference from our predefined view, we re-trained our model by adjusting the elevation angle of our default views with $\ang{20}$ in our synthetic training data. This new model is denoted by LAS-Diffusion$^\star$. \cref{tab:prosketch_chair} reports the performance of our method and other competing methods. We can see that LAS-Diffusion$^\star$ achieves the best performance. \cref{fig:prosketch} illustrates our results.


\begin{table}[t]
    \caption{Quantitative evaluations on the ProSketch dataset. The units of Sketch-CD, CD, EMD and Voxel-IOU are the same as in \cref{tab:ikea_sketch}.}
    \label{tab:prosketch_chair}
    \centering
    \resizebox{\columnwidth}{!}
    {

        \begin{tabular}{c|cc|ccc}
            \toprule
            \thead{Method}        & \thead{CLIPScore$\uparrow$} & \thead{Sketch-CD$\downarrow$} & \thead{CD$\downarrow$} & \thead{EMD$\downarrow$} & \thead{Voxel-IOU$\uparrow$} \\ \midrule
            Sketch2Model          & 86.52                       & 214.8                         & 105.0                  & 30.10                   & 12.76                       \\
            Sketch2Mesh           & 89.84                       & 43.29                         & 21.39                  & 17.13                   & 28.75                       \\
            SketchSampler         & N/A                         & N/A                           & 58.25                  & 24.04                   & 21.81                       \\
            SketchSampler$_m$     & 89.40                       & 59.88                         & 55.24                  & 24.47                   & 19.64                       \\
            LAS-Diffusion         & 93.36                       & 55.62                         & 26.04                  & 16.07                   & 33.43                       \\
            LAS-Diffusion$^\star$ & \textbf{93.70}              & \textbf{39.75}                & \textbf{19.56}         & \textbf{14.73}          & \textbf{34.97}              \\
            \bottomrule
        \end{tabular}
    }
\end{table}

\begin{figure}[t]
    \centering
    \includegraphics[width=0.95\linewidth]{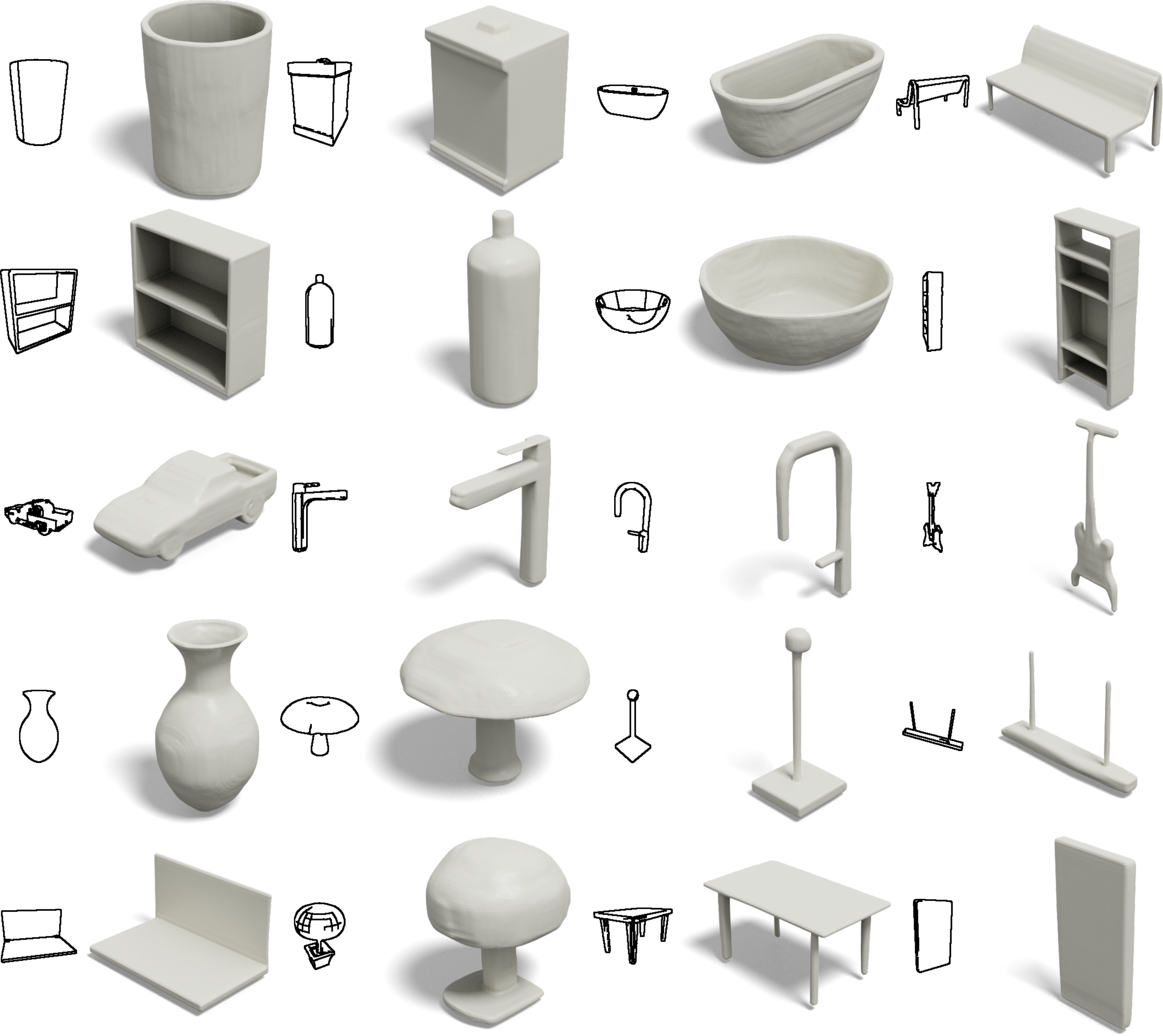}
    \caption{We trained our LAS-Diffusion model on the whole ShapeNetV1 dataset. Some randomly picked results are visualized. }
    \label{fig:shape55} 
\end{figure} 

\paragraph{Model scalability} Our LAS-Diffusion model is scalable to more diverse sketch data and random views. We trained a single LAS-Diffusion model on the whole ShapeNetV1 dataset, without restricting views to predefined views. During training, each shape is rendered from random views to generate sketches.  \cref{fig:shape55} visualizes the sketch-conditioned generative results by this model.

\begin{figure}[t]
    \centering
    \includegraphics[width=0.95\linewidth]{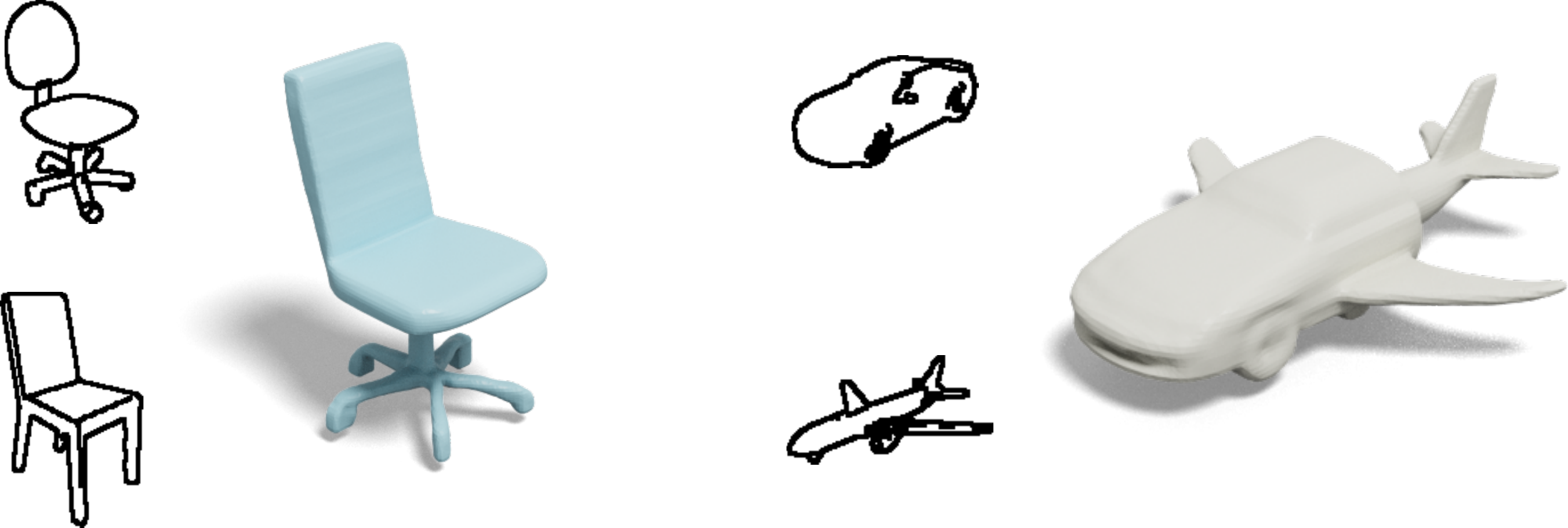}
    \caption{Shape generation via ViT feature manipulation. \textbf{Left}: The bottom half patch features of the swivel chair and the top half patch features of the four-legged chair are stitched together. \textbf{Right}: The left half-patch features of the car and the right half-patch features of the airplane are stitched together. In both cases,  novel and meaningful shapes are generated by LAS-Diffusion, without drawing new sketches. }
    \label{fig:vitedit} 
\end{figure}

\paragraph{Shape generation via ViT feature manipulation} Our model supports a new way to generate novel shapes by swapping ViT patch features of two existing sketches. For instance, we can replace the top-half patch features of a sketch with the bottom-half features of another sketch, to mimic a shape assembled by the top-half part of the first shape and the bottom-half of another shape. In \cref{fig:vitedit}, we present two novel shapes generated in this way. These interesting results indicate that ViT feature manipulation can be a novel and promising way of controlling shape generation.

\subsection{Ablation Studies} \label{subsec:ablation}
We designed two alternative attention mechanisms to replace our view-aware local attention.
\begin{enumerate}[leftmargin=*]\setlength\itemsep{1mm}
    \item[-] \emph{Global attention}. Instead of using local patch features, we directly
    use the global image feature to guide voxel feature learning. The global feature is from the classification token of the pre-trained ViT backbone. It is projected via MLP to align with the U-Net feature dimension, and associate with the U-Net feature vectors of the occupancy diffusion module at each level via element-wise multiplication.

    \item[-] \emph{View-agnostic attention}.
    We let the voxel features in the U-Net of the occupancy-diffusion module interact with all the image patch features via cross-attention, \ie the mask $\mathcal{M}$ in \cref{eq:crossattn} is \texttt{None}. In this way, no view information is required. The involved levels remain unchanged.
\end{enumerate}

\begin{figure}[t]
    \centering
    \begin{overpic}[width=0.9\linewidth]{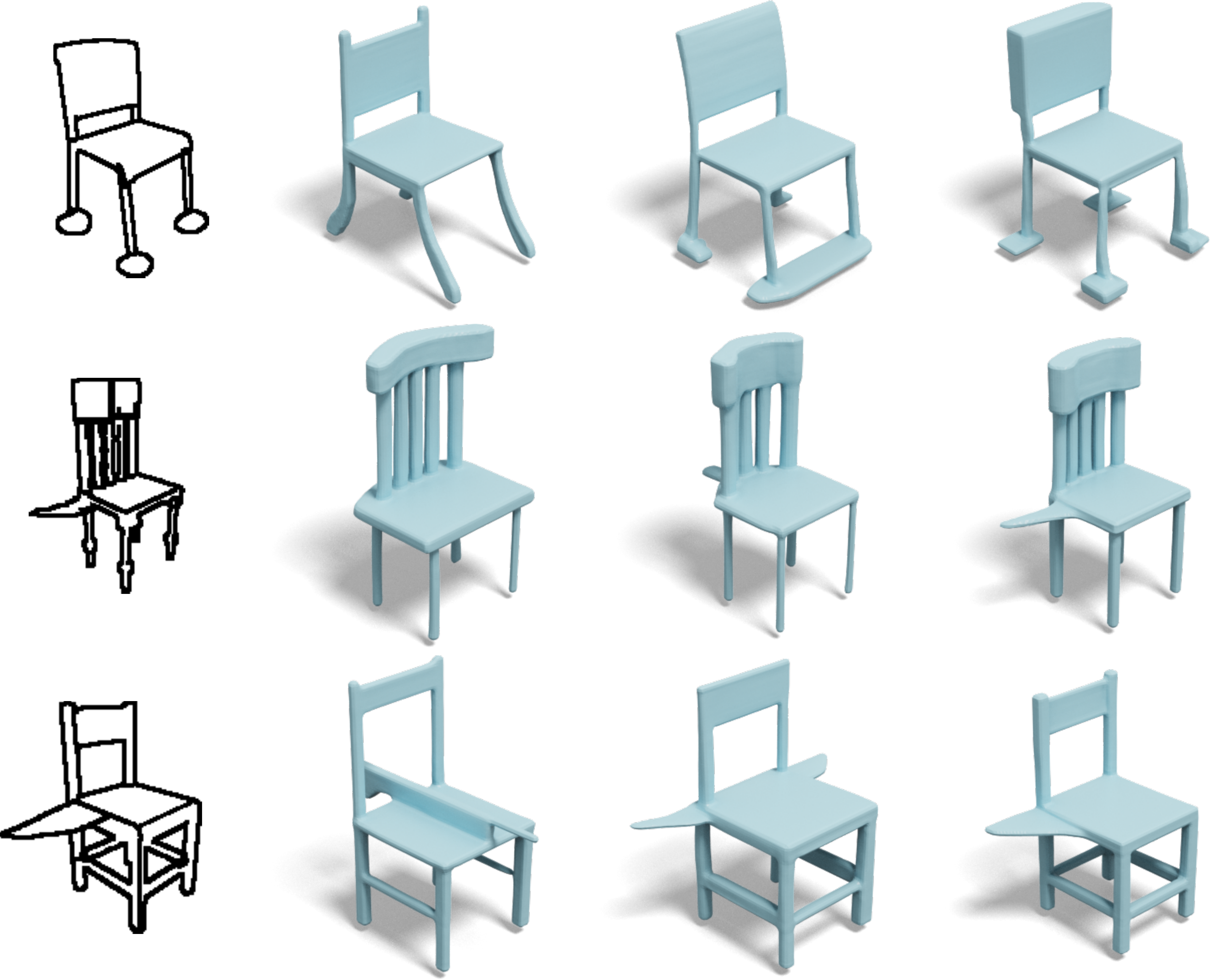}
        \put(32,-3){\scriptsize \textbf{global}}
        \put(56,-3){\scriptsize \textbf{view-agnostic}}
        \put(87,-3){\scriptsize \textbf{default}}
    \end{overpic}
    \caption{Ablation studies on different attention mechanisms. From \textbf{left} to \textbf{right}: global attention, view-agnostic attention,  and our default attention --- view-aware local attention. }
    \label{fig:ablation} 
\end{figure}

We trained LAS-Diffusion using the above attention mechanisms and found that: (1) \emph{Global attention} has very limited generalizability, and cannot process unseen shape structures; (2) \emph{View-agnostic attention} responds to sketch variations but often yields additional and wrong geometry, due to loss of local attention. \cref{fig:ablation} illustrates these issues.

\paragraph{Ablation study on neighborhood size} As seen in the above ablation study, view-agnostic attention does not yield satisfying results, because the attention region is too large and makes feature learning harder. As the local attention plays an important role, we examine how the neighborhood size  $d_\delta$ affects our model, by varying $d_\delta$ from the default $4\times \texttt{patch\_width}$ to $2 \times \texttt{patch\_width}$ and $6  \times \texttt{patch\_width}$ and retraining our model.  We found that the models with these small neighborhood sizes have similar performance.
Our default setting has the best CLIP scores on IKEA chairs: 96.63 ($2 \times \texttt{patch\_width}$), \textbf{96.92} ($4\times \texttt{patch\_width}$), and 96.67 ($6  \times \texttt{patch\_width}$).

\section{Category-conditioned Shape Generation} \label{sec:classcondition}
In this section, we conducted extensive experiments and evaluations on the task of category-conditioned shape generation.

\paragraph{Dataset} We also use the five shape categories from ShapeNetV1: \texttt{chair}, \texttt{car}, \texttt{airplane}, \texttt{table}, and \texttt{rifle}, to verify the generation capability and quality of LAS-Diffusion without sketch input. We follow the train/val/test split of \cite{chen2019learning}.

\paragraph{Training configurations}  We have two training configurations, depending on whether a single category is used for training.
\begin{enumerate}[leftmargin=*]\setlength\itemsep{1mm}
    \item[-]  \emph{Single-category generation}: we train our model in a single-category manner, \ie there are 5 LAS-Diffusion models in total. This per-category model is setup for a fair comparison with other existing 3D shape generative models.
    \item[-] \emph{Multi-category-conditioned generation}: we train a single LAS-Diffusion model on 5 categories to evaluate model scalability. We encode the class name, such as ``a chair'', by a pre-trained CLIP network~\cite{radford2021learning}. We associate the CLIP feature with the U-Net feature vectors of the occupancy diffusion module, as conditions, similar to \emph{global attention} in \cref{subsec:ablation}. We found empirically that it is not necessary to add CLIP features into the SDF-diffusion stage.
\end{enumerate}

\paragraph{Training details}
For both configurations, we trained the occupancy-diffusion module using AdamW optimizer with a fixed learning rate of $10^{-4}$ over $4000$ epochs; and reused the trained SDF-diffusion module from~\cref{sec:sketchcondition}.

\begin{figure*}[t]
    \centering
    \includegraphics[width=0.95\linewidth]{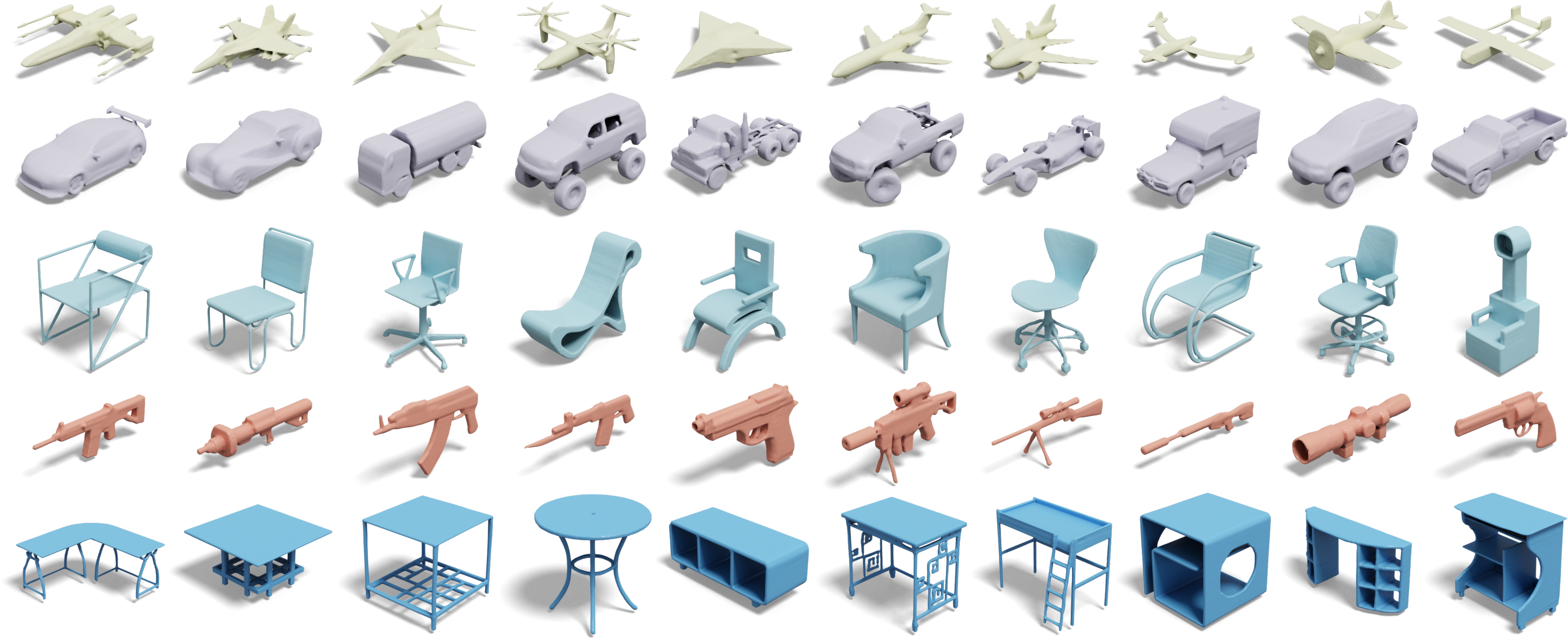}
    \caption{Single-category generation results. Our models were trained on five shape categories: \texttt{airplane}, \texttt{car}, \texttt{chair}, \texttt{rifle}, and \texttt{table}. We selected ten generated shapes from each category to demonstrate shape diversity and high-quality geometry.}
    \label{fig:classcondition} 
\end{figure*} 

\paragraph{Evaluation metric} To evaluate the quality and diversity of generated shapes, we adopt the metric proposed by
\cite{zheng2022sdfstylegan}: \emph{shading-image-based FID}, which avoids the drawback of existing metrics built on light-field-distance (LFD) or 3D mesh distances. To compute this metric, each generated shape was rendered from 20 uniformly distributed views, and its shading images are used to compute FID scores, on the rendered image set of the original training dataset.
The metric formula is defined as follows.
\begin{equation}
    FID = \frac{1}{20}\left[\sum_{i=1}^{20} \|\mu_{g}^i - \mu_{r}^i\|^2 + \operatorname{Tr}\left(\Sigma_{g}^i + \Sigma_{r}^i - 2 \left(\Sigma_{r}^i\Sigma_{g}^i\right)^{1/2}\right)\right],
\end{equation}
where $g$ and $r$ denote the features of the generated data set and the training set, $\mu^i$, $\Sigma^i$ denote the mean and covariance matrices of the shading images rendered from the $i$-th view, respectively. A lower FID indicates better generation quality and diversity.

\paragraph{Evaluation and comparisons} \cref{fig:classcondition} illustrates high-quality and diverse generation results by our single-category LAS-Diffusion model. The results show that our model can generate structurally complex shapes with fine geometry.  More uncurated generated results including intermediate occupancy generation are provided in the supplemental material.

We compared our approach with four representative 3D generation models, including two GAN-based models: IM-GAN~\cite{chen2019learning} and SDF-StyleGAN~\cite{zheng2022sdfstylegan},  a 3D diffusion model: Wavelet-Diffusion~\cite{hui2022neural} and an autoregressive model: 3DILG~\cite{zhang20223dilg}. We use their pre-trained models for evaluation. Except for 3DILG which was trained on all ShapeNetV2 data and our multi-category-conditioned model, other methods were trained on a single category. Here, IM-GAN and Wavelet-Diffusion use $256^3$ occupancy and SDF fields as ground truth for training, respectively.

\begin{table}[t]
    \caption{Quantitative comparison with different methods. The reported numbers are shading-image-based FID scores (lower is better). LAS-Diffusion$^\dagger$ and LAS-Diffusion$^\ddagger$ denote the single-category models and multi-category-conditioned model, respectively. Note that Wavelet-Diffusion was trained on 3 categories only.
    }
    \label{tab:fid}
    \centering
    \resizebox{1\columnwidth}{!}
    {
        \begin{tabular}{c|ccccc}
            \toprule
            \thead{Method}       & \thead{Chair}  & \thead{Airplane} & \thead{Car}    & \thead{Table}  & \thead{Rifle}  \\
            \midrule
            IM-GAN               & 63.42          & 74.57            & 141.2          & 51.70          & 103.3          \\

            SDF-StyleGAN         & 36.48          & 65.77            & 97.99          & 39.03          & 64.86          \\
            Wavelet-Diffusion    & 28.64          & 35.05            & N/A            & 30.27          & N/A            \\
            LAS-Diffusion$^\dagger$ & \textbf{20.45} & \textbf{32.71}   & \textbf{80.55} & \textbf{17.25} & \textbf{44.93} \\
            \midrule
            3DILG                & 31.64          & 54.38            & 164.15         & 54.13          & 77.74          \\
            LAS-Diffusion$^\ddagger$   & 21.55          & 43.08            & 86.34          & 17.41          & 70.39          \\
            \bottomrule
        \end{tabular}
    }
\end{table} 

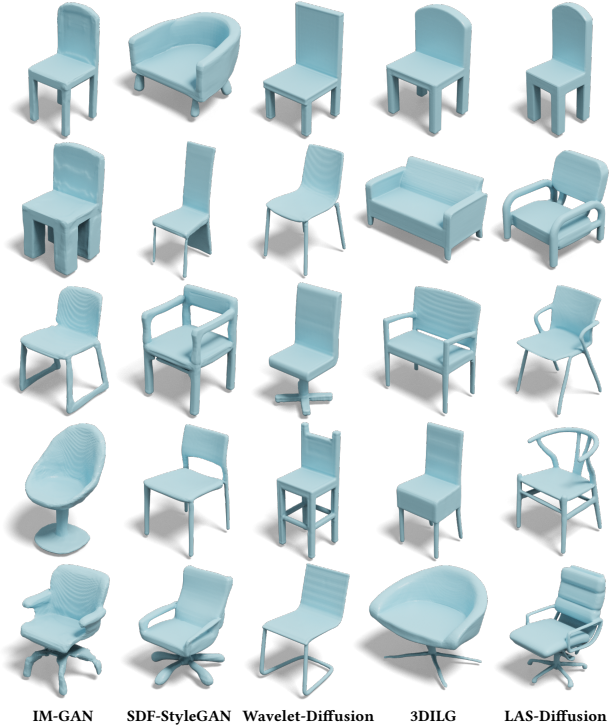
\begin{figure}[t]
    \centering
    \begin{overpic}[width=0.95\linewidth]
        {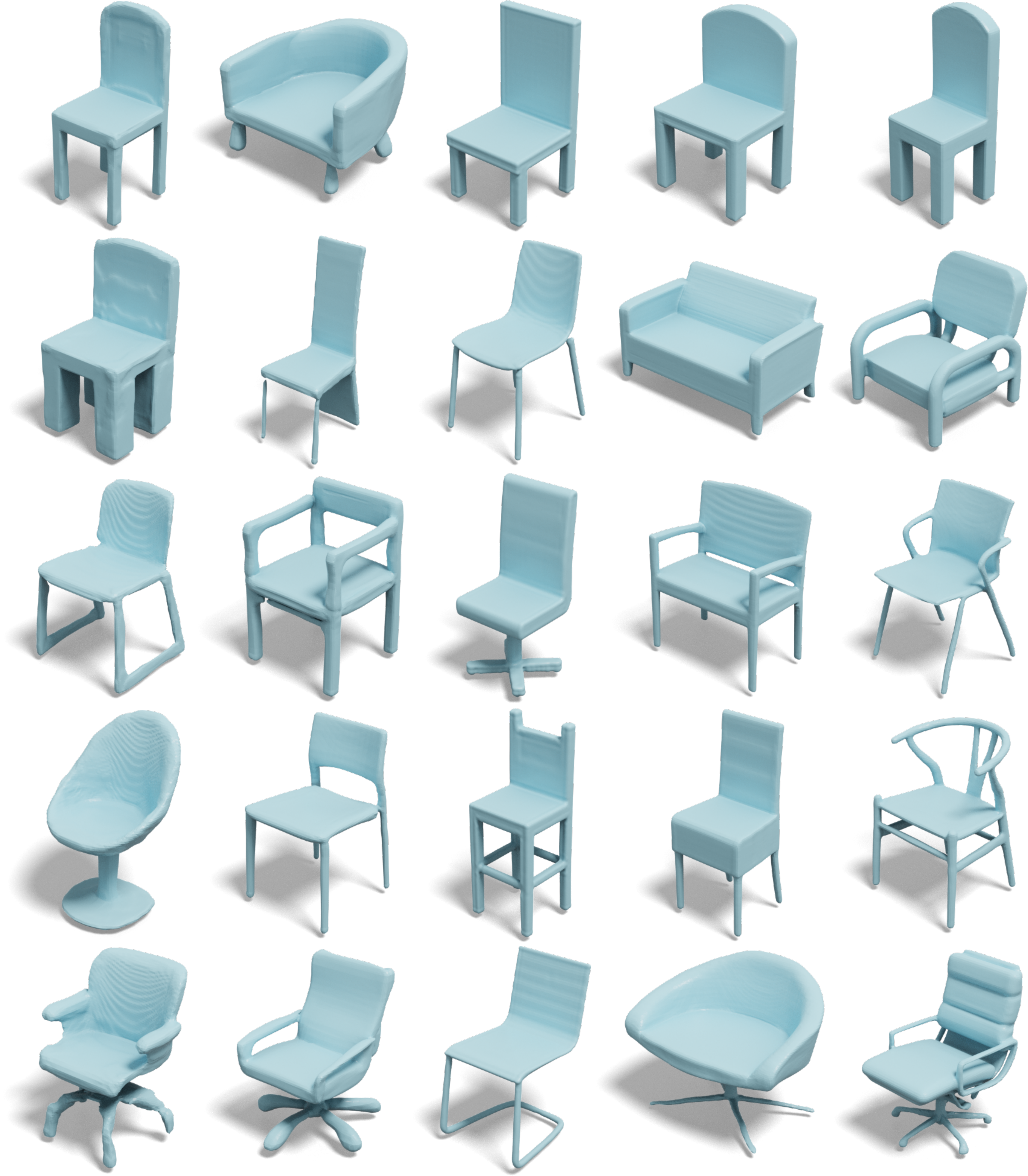}
        \put(5,-3){\scriptsize \textbf{IM-GAN}}
        \put(18.5,-3){\scriptsize \textbf{SDF-StyleGAN}}
        \put(35, -3){\scriptsize \textbf{Wavelet-Diffusion} }
        \put(59,-3){\scriptsize \textbf{3DILG}}
        \put(72.5,-3){\scriptsize \textbf{LAS-Diffusion}}
    \end{overpic}
    \caption{Randomly selected chairs generated by different methods.}
    \label{fig:visualcompare} 
\end{figure}

\cref{tab:fid} reports the FID scores of all methods. We conclude that: (1) our single-category LAS-Diffusion outperforms other methods in all five categories;  (2) our multi-category-conditioned LAS-Diffusion is slightly worse than its unconditioned version, but still performs better than other methods, except for \texttt{airplane} (Wavelet-Diffusion) and \texttt{rifle} (SDF-StyleGAN).  The comparison with 3DILG is for reference only, as their training data are not exactly the same.
In \cref{fig:visualcompare}, we visualize some chairs generated by different methods. We can see that Wavelet-Diffusion, 3DILG, and our LAS-Diffusion are visually comparable and possess a more faithful geometry than IM-GAN and SDF-StyleGAN, furthermore, the meshes generated by our method have less bumpy geometry than others.

Following \cite{hui2022neural}, we also adopt the COV, MMD, and 1-NNA metrics~\cite{achlioptas2018learning,yang2019pointflow}  based on the Chamfer distance (CD) and the Earth mover's distance (EMD) on the sampled points to access the fidelity, coverage, and diversity of generative models. Lower MMD, higher COV, 1-NNA that has a smaller difference to \SI{50}{\percent}, mean better quality. We report these metrics in the \texttt{chair} category in \cref{tab:traiditional} for IM-GAN, SDF-StyleGAN, Wavelet-Diffusion, and our single-category model. 2048 points were sampled on each mesh uniformly to perform the evaluation. Wavelet-Diffusion and our method are comparable on MMD and 1-NNA, and our method attains better COV(EMD) than other methods.

\begin{table}[t]
    \caption{Additional metric evaluations on \texttt{chair} category. The units of CD and EMD are $10^{-3}$ and $10^{-2}$, respectively. LAS-Diffusion$^\dagger$ denotes the single-category model.}
    \label{tab:traiditional}
    \resizebox{1\columnwidth}{!}
    {
        \begin{tabular}{c|cccccc}
            \toprule
            \multirow{2}{*}{\thead{Method}} & \multicolumn{2}{c}{\thead{COV(\%)$\uparrow$}} & \multicolumn{2}{c}{\thead{MMD$\downarrow$}} & \multicolumn{2}{c}{\thead{1-NNA(\%)$\downarrow$}}                                                    \\
                                            & \thead{CD}                      & \thead{EMD}                     & \thead{CD}                        & \thead{EMD}    & \thead{CD}     & \thead{EMD}    \\ \midrule
            IM-GAN                          & \textbf{57.30}                  & 49.48                           & \textbf{13.12}                    & 17.70          & 62.24          & 69.32          \\
            SDF-StyleGAN                    & 52.36                           & 48.89                           & 14.97                             & 18.10          & 65.38          & 69.06          \\
            Wavelet-Diffusion               & 52.88                           & 47.64                           & 13.37                             & \textbf{17.33} & \textbf{61.14} & 66.92          \\
            LAS-Diffusion$^\dagger$            & 53.76                           & \textbf{52.43}                  & 13.79                             & 17.45          & 64.53          & \textbf{65.15} \\
            \bottomrule
        \end{tabular}
    } 
\end{table}

\begin{figure}[t]
    \centering
    \begin{overpic}[width=0.75\columnwidth]{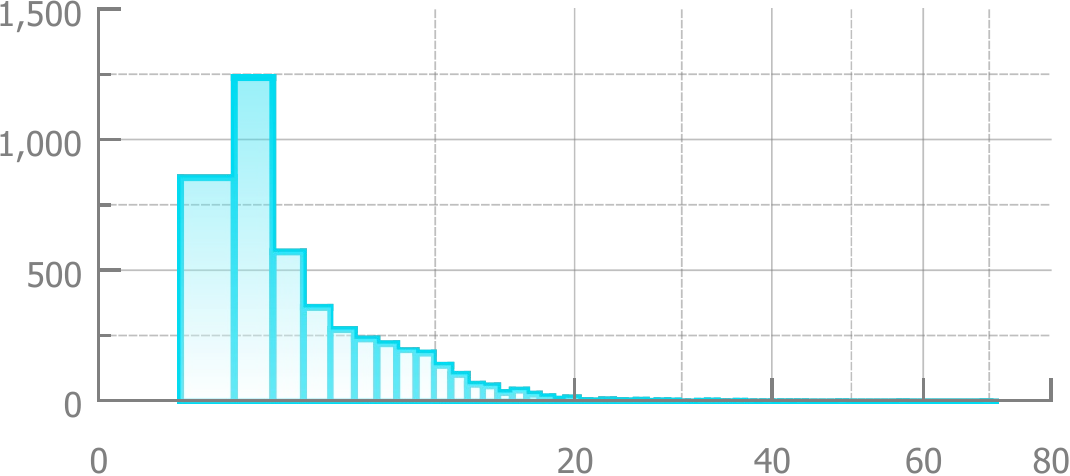}
        \put(100, 8){\scriptsize CD}
    \end{overpic}
    \\
    \vspace{2mm}
    \begin{overpic}[width=0.8\linewidth]{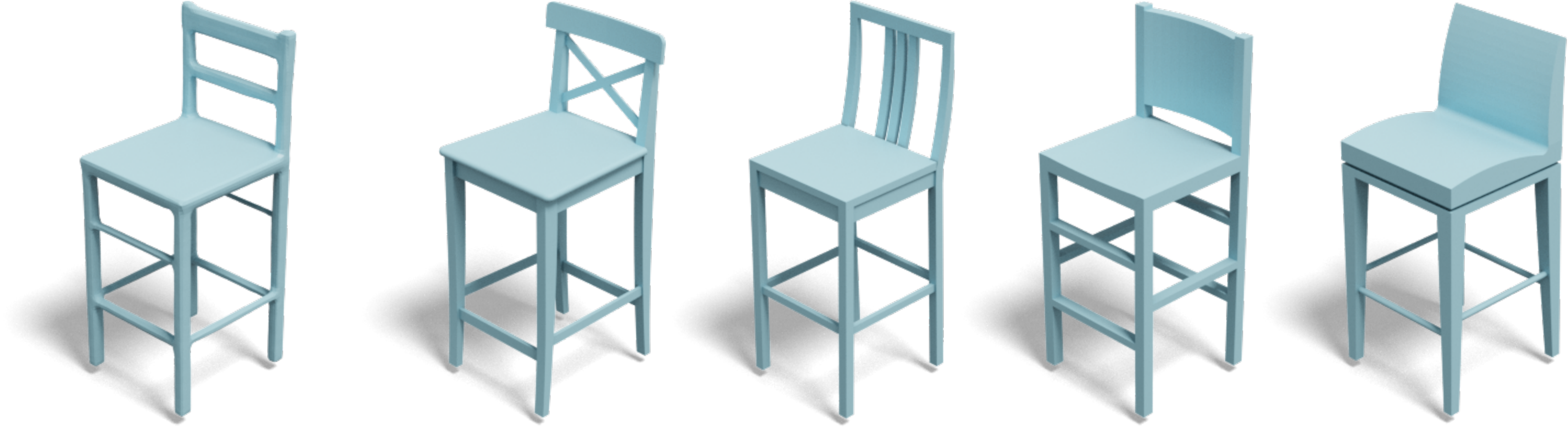}
        \put(22,0){
            \begin{tikzpicture}
                \draw (0,0) -- (0,1.87);
            \end{tikzpicture}
        }
    \end{overpic}
    \caption{\textbf{Top}: The histogram on the distribution of the Chamfer distance (CD) between the generated chairs and the training dataset. Here, we use SquareRoot binning for histogram drawing.  \textbf{Bottom}: The four nearest shapes (right) retrieved from the training dataset according to their Chamfer distance to the generated chair (left).}
    \label{fig:retrieval} 
\end{figure}
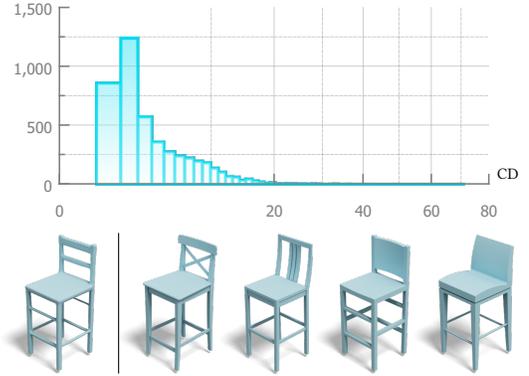 

\paragraph{Shape diversity} We also evaluate the model diversity of our method on \texttt{chair} category, by computing the histogram of Chamfer distance between the generated shapes and the training data. \cref{fig:retrieval}-top shows the histogram whose x-axis is Chamfer distance ($\times 10^3$).  The histogram reveals that most generated shapes are different from the training set.  \cref{fig:retrieval}-bottom presents a test case: for the generated chair (left), we retrieve the four most similar chairs (right) from the training dataset based on the Chamfer distance. We can see that the generated chair has a novel structure.

\begin{figure}[t]
    \centering
    \includegraphics[width=0.95\linewidth]{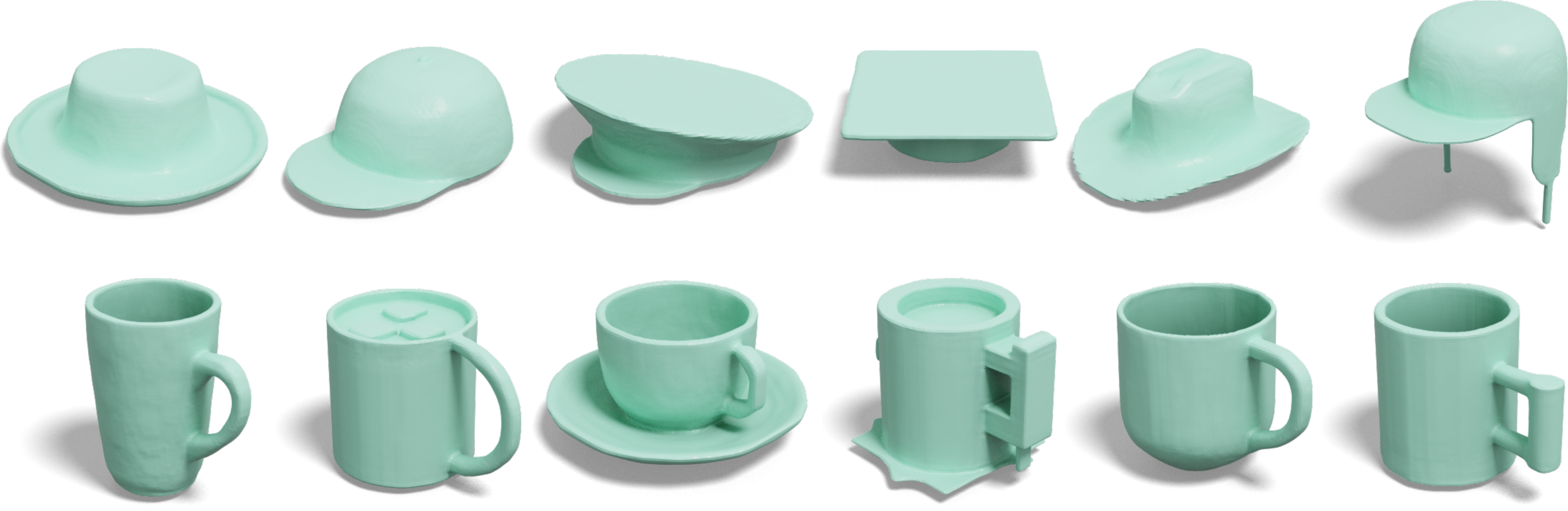}
    \caption{Generation results on small datasets. \textbf{Top}: generated results on \texttt{cap} category. \textbf{Bottom}: generated results on  \texttt{mug} category.}
    \label{fig:small_set} 
\end{figure} 

\paragraph{Small datasets} We also tested the capability of our single-category LAS-Diffusion on ShapeNet categories that have a small number of objects. We chose \texttt{cap} category (56 objects) as well as \texttt{mug} category (214 objects) to train the occupancy-diffusion module and reused the trained SDF-diffusion module. \cref{fig:small_set} shows that the shapes generated by LAS-Diffusion are plausible with good quality.

\section{Conclusion and Perspectives} \label{sec:conclusion}
We present a diffusion-based generative technique to synthesize plausible 3D shapes. Our view-aware local attention mechanism is well integrated with our two-stage diffusion model and offers greater controllability and generalizability than existing shape synthesis techniques. As this mechanism is simple and flexible, there is no difficulty in extending it to color images, depth images, and even 3D point cloud inputs.  We believe that it will be a powerful module for multimodal-conditioned content generation.

\begin{wrapfigure}[7]{r}{0.19\textwidth}
    \centering
    \includegraphics[width=0.19\textwidth]{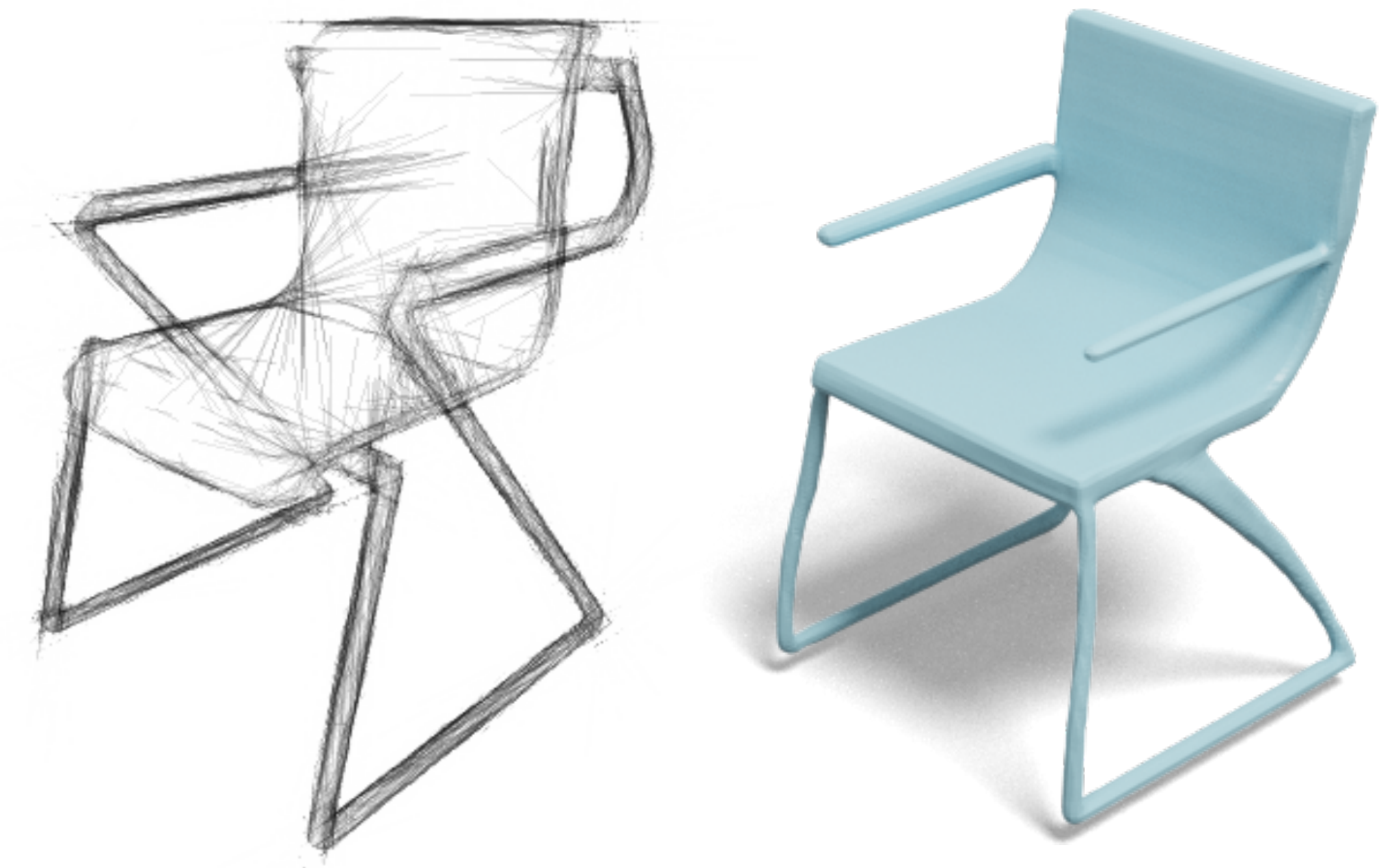}
    \vspace{-4mm}
\end{wrapfigure}

\paragraph{Limitations}
Currently, our model is trained on synthetic data only. The sketch style is tied to our rendering pipeline; therefore, our trained model is not well adapted for sketches with highly distorted lines, oversketches, or seriously inconsistent perspectives. An example is shown in the right inset, in which our model fails to generate the chair arm structure for the input with oversketches. We believe that this issue can be overcome by using more real sketches and paired 3D shapes for training.

In the future, we would like to explore the following directions.

\paragraph{Shape appearance} Currently our work focuses on shape geometry only and does not provide vivid shape appearances. As 2D sketches do not contain rich appearance information, we plan to leverage both 2D sketches and language descriptions to generate geometry-compatible and plausible shape appearances.

\paragraph{Multi-view sketches} Single-view sketches do not convey the complete idea of designers. We will study how to utilize multi-view sketches in our model and provide a convenient user interface to assist 3D design.

\bibliographystyle{ACM-Reference-Format}
\bibliography{src/reference}
\end{document}